\documentclass[runningheads]{llncs}
\usepackage[misc,geometry]{ifsym}
\usepackage{amssymb}
\usepackage{graphicx}
\usepackage{romannum}
\usepackage{multirow}
\usepackage{rotating}
\usepackage{comment}
\usepackage{subfig}
\usepackage{xcolor}
\usepackage[export]{adjustbox}
\newcommand{\layer}[1]{\small{\texttt{#1}}}
\def\eg{\textit{e.g.\,}}
\def\ie{\textit{i.e.\,}}
\def\etc{\textit{etc.\,}}
\def\wrt{\textit{w.r.t.\,}}

\def\sauvola{\textit{Sauvola}\,}
\def\snet{\texttt{SauvolaNet}\,}
\usepackage{amssymb}
\usepackage{pifont}
\usepackage{hyperref}
\hypersetup{
    colorlinks=true,
    linkcolor=magenta,
    filecolor=blue,      
    urlcolor=magenta,
}
\usepackage{amsmath}
\usepackage{mathtools}
\DeclareMathOperator{\argmax}{argmax}

\DeclarePairedDelimiter\floor{\lfloor}{\rfloor}
\def\opencv{\textit{OpenCV}\,}
\def\pythreshold{\textit{Pythreshold}\,}
\def\skimage{\textit{Scikit-Image}\,}
\newcommand{\cmark}{\ding{51}}%
\newcommand{\xmark}{\ding{55}}%
\usepackage{color, colortbl}
\usepackage{caption}
\begin{document}
\pagenumbering{arabic}
\title{\texttt{\textbf{SauvolaNet}}: Learning Adaptive Sauvola Network for Degraded Document Binarization}
%
%

\author{Deng Li\inst{1} \and
Yue Wu\inst{2} \and
Yicong Zhou\inst{1}(\Letter) \orcidID{0000-0002-4487-6384}}
\authorrunning{Deng Li et al.}
%
\institute{ University of Macau, Department of Computer and Information Science, Macau, China
\email{\{mb85511,yicongzhou\}@um.edu.com}\\\and
Amazon Alexa Natural Understanding, Manhattan Beach, CA, USA\\
\email{wuayue@amazon.com}}
\titlerunning{SauvolaNet: Learning Adaptive Sauvola Network}
\maketitle              

\begin{abstract}
Inspired by the classic \sauvola{} local image thresholding approach, we systematically study it from the deep neural network (DNN) perspective and propose a new solution called \snet{} for degraded document binarization (DDB). It is composed of three explainable modules, namely, Multi-Window Sauvola (MWS), Pixelwise Window Attention (PWA), and Adaptive Sauolva Threshold (AST). The MWS module honestly reflects the classic \sauvola{} but with trainable parameters and multi-window settings. The PWA module estimates the preferred window sizes for each pixel location. The AST module further consolidates the outputs from MWS and PWA and predicts the final adaptive threshold for each pixel location. As a result, \snet{} becomes end-to-end trainable and significantly reduces the number of required network parameters to 40K -- it is only 1\% of \texttt{MobileNetV2}. In the meantime, it achieves the State-of-The-Art (SoTA) performance for the DDB task -- \snet{} is at least comparable to, if not better than, SoTA binarization solutions in our extensive studies on the 13 public document binarization datasets. Our source code is available at \url{https://github.com/Leedeng/SauvolaNet}.

\end{abstract}
\footnotetext[1]{This work was done prior to Amazon involvement of the authors.}
\footnotetext[2]{This work was funded by The Science and Technology Development Fund, Macau SAR (File no. 189/2017/A3), and by University of Macau (File no. MYRG2018-00136-FST).}

\section{Introduction}\label{sec:introduction}
Document binarization typically refers to the process of taking a gray-scale image and converting it to black-and-white. Formally, it seeks a decision function $f_{\textrm{binarize}}(\cdot)$ for a document image $\mathbf{D} $ of width $W$ and height $H$, such that the resulting image $\hat{\mathbf{B}}$ of the same size only contains binary values while the overall document readability is at least maintained if not enhanced.
\begin{equation}
\hat{\mathbf{B}}=f_{\textrm{binarize}}(\mathbf{D})
\end{equation}
Document binarization plays a crucial role in many document analysis and recognition tasks. It is the prerequisite for many low-level tasks like connected component analysis, maximally stable extremal regions, and high-level tasks like text line detection, word spotting, and optical character recognition (OCR). 

Instead of directly constructing the decision function $f_{\textrm{binarize}}(\cdot)$, classic binarization algorithms~\cite{otsu1979threshold,sauvola2000adaptive} typically first construct an auxiliary function $g(\cdot)$ to estimate the required thresholds $\mathbf{T}$ as follows.
\begin{equation}\label{eq:g}
    \mathbf{T} = g_{\textrm{classic}}(\mathbf{D})
\end{equation}
In global thresholding approaches~\cite{otsu1979threshold}, this threshold $\mathbf{T}$ is a scalar, \ie all pixel locations use the same threshold value. In contrast, this threshold $\mathbf{T}$ is a tensor with different values for different pixel locations in local thresholding approache~\cite{sauvola2000adaptive}. Regardless of global or local thresholding, the actual binarization decision function can be written as
\begin{equation}\label{eq:fg}
\hat{\mathbf{B}}_{\textrm{classic}}=f_{\textrm{classic}}(\mathbf{D}) = th(\mathbf{D}, \mathbf{T}) = th(\mathbf{D}, g_\textrm{classic}(\mathbf{D})) \end{equation}
where $th(x,y)$ is the simple thresholding function and the binary state for a pixel located at $i$-th row and $j$-th column is determined as in Eq.~\eqref{eq:bij}.

\begin{equation}\label{eq:bij}
	\hat{B}_{\textrm{classic}}[i,j]=th(D[i,j], T[i,j]) = \begin{cases}
	+1, &\textrm{if\,} D[i,j]\geq T[i,j]\\
	-1, &\textrm{otherwise} 	
		   \end{cases}
\end{equation}

Classic binarization algorithms are very efficient in general because of using simple heuristics like intensity histogram~\cite{otsu1979threshold} and local contrast histogram~\cite{su2012robust}. The speed of classic binarization algorithms typical of the millisecond level, even on a mediocre CPU. However, simple heuristics also means that they are sensitive to potential variations~\cite{su2012robust} (image noise, illumination, bleed-through, paper materials, \etc), especially when the relied heuristics fail to hold. In order to improve the binarization robustness, data-driven approaches like~\cite{wu2016learning} learn the decision function $f_{\textrm{binarize}}(\cdot)$ from data rather than heuristics. However, these approaches typically achieve better robustness by using much more complicated features, and thus work relatively slow in practice, \eg on the second level ~\cite{wu2016learning}.  

\begin{figure}[!t]
    \centering
    \includegraphics[width=.85\linewidth]{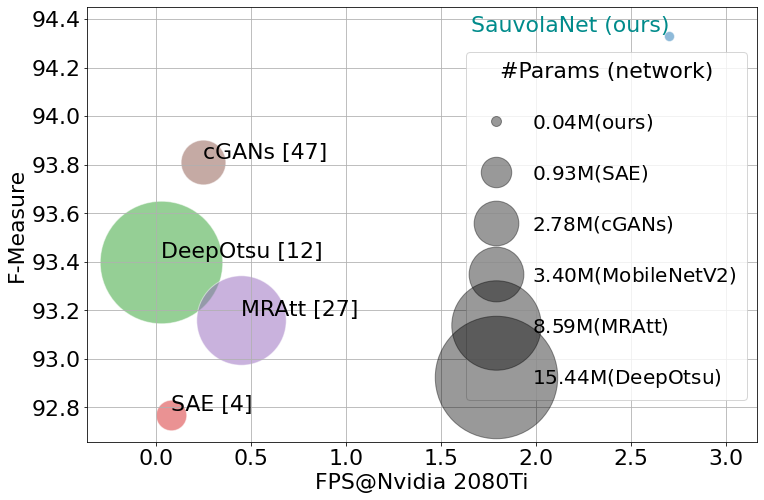}
    \caption{Comparisons of the SoTA DNN-based methods on the DIBCO 2011 dataset (average resolution 574$\times$1104). The {\textcolor[RGB]{0,139,139}{\snet{}}} is of 1\% of \texttt{MobileNetV2}'s parameter size, while attaining superior performance in terms of both speed and F-Measure.}
    \label{fig:blobPlot}
\end{figure}

Like in many computer vision and image processing fields, the deep learning-based approaches outperform the classic approaches by a large margin in degraded document binarization tasks. The state-of-the-art (SoTA) binarization approaches are now all based on deep neural networks (DNN)~\cite{pratikakis2017icdar2017,Pratikakis2019icdar}. Most of SoTA document binarization approaches~\cite{calvo2019selectional,peng2019document,vo2018binarization} treat the degraded binarization task as a binary semantic segmentation task (namely, foreground and background classes) or a sequence-to-sequence learning task~\cite{afzal2015document}, both of which can effectively learn $f_{\textrm{binarize}}(\cdot)$ as a DNN from data. 

Recent efforts~\cite{calvo2019selectional,peng2019document,he2019deepotsu,zhao2019document,de2020document,souibgui2020gan,vo2018binarization} focus more on improving robustness and generalizability. In particular, the SAE approach~\cite{calvo2019selectional} suggests estimating the pixel memberships not from a DNN's raw output but the DNN's activation map, and thus generalizes well even for out-of-domain samples with a weak activation map. The MRAtt approach ~\cite{peng2019document} further improves the attention mechanism in multi-resolution analysis and enhances the DNN's robustness to font sizes. DSN~\cite{vo2018binarization} apply multi-scale architecture to predict foreground pixel at multi-features levels. The {DeepOtsu} method~\cite{he2019deepotsu} learns a DNN that iteratively enhances a degraded image to a uniform image and then binarized via the classic Otsu approach. Finally, generative adversarial networks (GAN) based approaches like cGANs~\cite{zhao2019document} and DD-GAN~\cite{de2020document} rely on the adversarial training to improve the model's robustness against local noises by penalizing those problematic local pixel locations that the discriminator uses in differentiating real and fake results. 

As one may notice, both classic and deep binarization approaches have pros and cons: 1) the classic binarization approaches are extremely fast, while the DNN solutions are not; 2) the DNN solutions can be end-to-end trainable, while the classic approaches can not. In this paper, we propose a novel document binarization solution called \snet{} -- it is an end-to-end trainable DNN solution but analogous to a multi-window \sauvola{} algorithm. More precisely, we re-implement the \sauvola{} idea as an algorithmic DNN layer, which helps \snet{} attain highly effective feature representations at an extremely low cost -- only two \sauvola{} parameters are needed. We also introduce an attention mechanism to automatically estimate the required \sauvola{} window sizes for each pixel location and thus could effectively and efficiently estimate the \sauvola{} threshold. In this way, the \snet{} significantly reduces the total number of DNN parameters to 40K, only 1\% of the \texttt{MobileNetV2}, while attaining comparable performance of SoTA on public DIBCO datasets. Fig.~\ref{fig:blobPlot} gives the high-level comparisons of the proposed \snet{} to the SoTA DNN solutions.

The rest of the paper is organized as follows: Sec.~\ref{sec:related} briefly reviews the classic \sauvola{} method and its variants; Sec.~\ref{sec:method} proposes the \snet solution for degraded document binarization; Sec.~\ref{sec:experiment2} presents \sauvola{} ablation studies results and comparisons to SoTA methods; and we conclude the paper in Sec.~\ref{sec:conclusion}.

\section{Related \sauvola Approaches}\label{sec:related}

The \sauvola{} binarization algorithm~\cite{sauvola2000adaptive} is widely used in main stream image and document processing libraries and systems like \texttt{OpenCV}~\footnote{https://docs.opencv.org/4.5.1/} and \texttt{Scikit-Image}~\footnote{https://scikit-image.org/}. As aforementioned, it constructs the binarization decision function \eqref{eq:fg} via the auxiliary threshold estimation function $g_{\textrm{Sauvola}}$, which has three hyper-parameters, namely, 1) $w$: the square window size (typically an odd positive integer~\cite{cheriet2013learning}) for computing local intensity statistics; 2) $k$: the user estimated level of document degradation; and 3) $r$: the dynamic range of input image's intensity variation. 
\begin{equation}\label{eq:g_sauvola}
\mathbf{T}_{\textrm{Sauvola}} = g_{\textrm{Sauvola} | \theta}(\mathbf{D}).
\end{equation}
where $\theta\!\!=\!\!\{w, k, r\}$ indicates the used hyper-parameters. Each local threshold is computed \wrt the 1st- and 2nd-order intensity statistics as shown in Eq.~\eqref{eq:sauvola},
\begin{equation}\label{eq:sauvola}
 T_{\textrm{Sauvola}|\theta}[i,j] = \mu[i,j] \cdot \left(1 + k \cdot \left(\frac{\sigma[i,j]}{r}-1\right)\right)
\end{equation}
where $\mu[i,j]$ and $\sigma[i,j]$ respectively indicate the mean and standard deviation of intensity values within the local window as follows.
\begin{equation}\label{eq:mu}
    \mu[i,j] = \sum_{\delta_i=-\floor{w/2}}^{\floor{w/2}} \sum_{\delta_j=-\floor{w/2}}^{\floor{w/2}} {D[i+\delta_i, j+\delta_j] \over w^2}
\end{equation}
\begin{equation}\label{eq:sigma}
    \sigma^2[i,j]= \sum_{\delta_i=-\floor{w/2}}^{\floor{w/2}} \sum_{\delta_j=-\floor{w/2}}^{\floor{w/2}} {\left(D[i+\delta_i, j+\delta_j]-\mu[i,j]\right)^2 \over w^2}
\end{equation}

It is well known that heuristic binarization methods with hyper-parameters could rarely achieve their upper-bound performance unless the method hyper-parameters are individually tuned for each input document image~\cite{kaur2020modified}, and this is also the main pain point of \sauvola{} approach.

\begin{table}[!h]
    \centering\scriptsize
    \caption{Comparisons of various \sauvola{} document binarization approaches.}
    \begin{tabular}{c|r|c|c|c|r|rr}\hline\hline
    \textbf{Related} & \textbf{End-to-End} & \multicolumn{3}{c|}{\textbf{\underline{\sauvola{} Params.}}} & \textbf{\#Used} &\multicolumn{2}{c}{\textbf{\underline{Without}}} \\
    \textbf{Work} & \textbf{Trainable}? & $w$ & $k$ & $r$ & \textbf{Scales} & \textbf{Preproc}? & \textbf{Postproc}? \\\hline
        \sauvola{}~\cite{sauvola2000adaptive} & \xmark & \multicolumn{3}{c|}{user specified} & single & \cmark & \cmark\\\hline
        \cite{moghaddam2010multi} & \xmark & auto & user specified & auto & multiple & \xmark & \xmark \\\cline{3-6}
         \cite{lazzara2014efficient} & \xmark & \multicolumn{3}{c|}{fixed} & multiple & \cmark & \cmark\\\cline{3-6}
         \cite{hadjadj2016isauvola} & \xmark & \multicolumn{3}{c|}{fixed} & single & \xmark & \xmark\\\cline{3-6}
         \cite{kaur2020modified} & \xmark & auto & fixed & fixed & multiple & \xmark & \cmark \\\hline
         \snet & \cmark & auto & learned & learned & multiple & \cmark & \cmark \\
         \hline\hline
    \end{tabular}
    \label{tab:related_works}
\end{table}

Many efforts have been made to mitigate this pain point. For example, 
\cite{moghaddam2010multi} introduces a multi-grid \sauvola{} variant that analyzes multiple scales in the recursive way;
\cite{lazzara2014efficient} proposes a hyper-parameter free multi-scale binarization solution called {Sauvola MS~\cite{calvo2019selectional}} by combining \sauvola{} results of a fixed set of window sizes, each with its own empirical $k$ and $r$ values; 
\cite{hadjadj2016isauvola} improves the classic \sauvola{} by using contrast information obtained from pre-processing to refine \sauvola{}'s binarization; 
\cite{kaur2020modified} estimates the required window size $w$ in \sauvola{} by using the stroke width transform matrix. Table~\ref{tab:related_works} compares these approaches with the proposed \snet{}, and it is clear that only \snet{} is end-to-end trainable.

\section{The \texttt{SauvolaNet} Solution}\label{sec:method}

\definecolor{train}{RGB}{204,255,255}
\definecolor{test}{RGB}{204,229,255}

\begin{figure}[!h]
    \centering
    \begin{tabular}{cc}
    \multicolumn{2}{c}{\includegraphics[trim=0 35cm 0 1cm, clip, width=\linewidth]{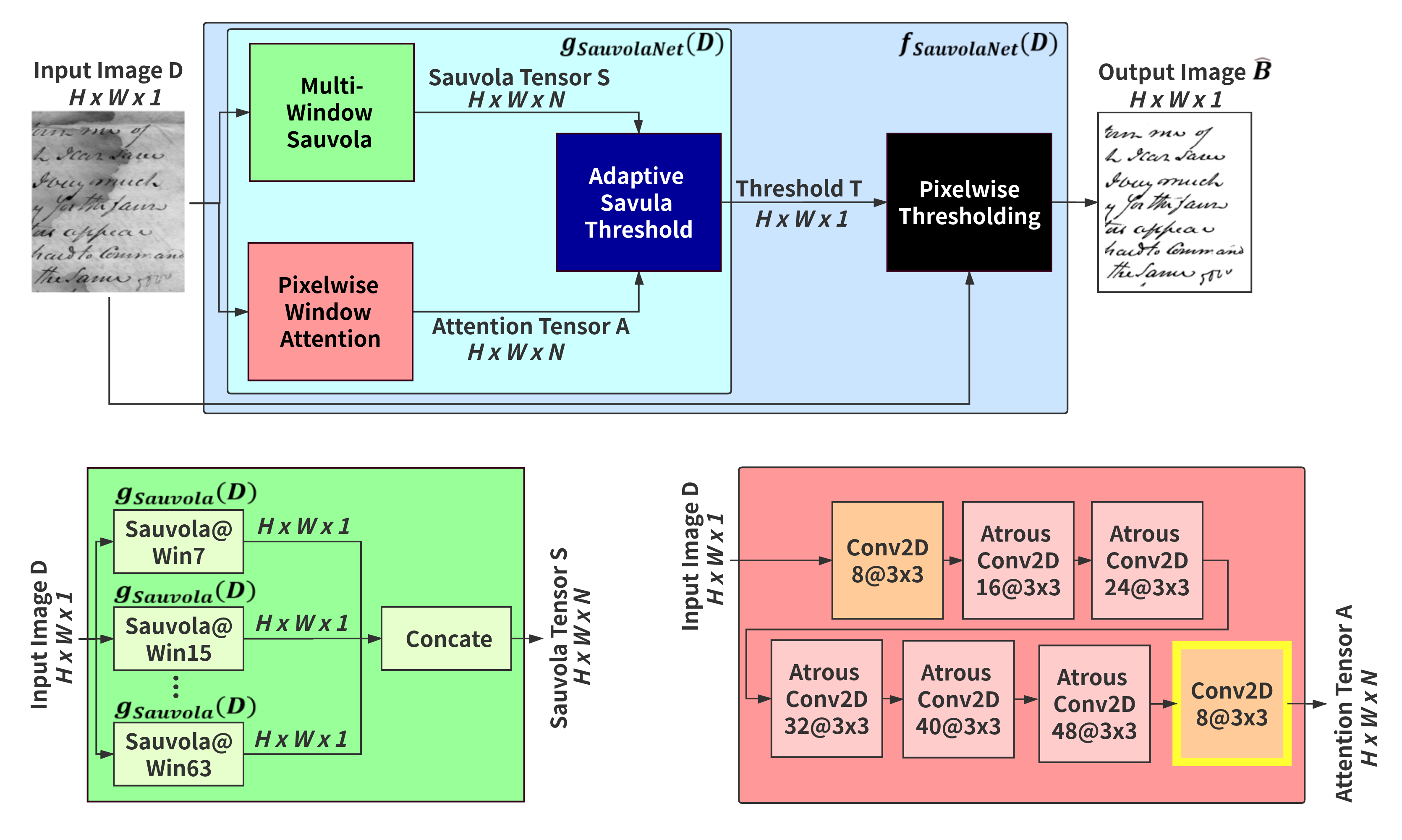}}\\
    \multicolumn{2}{c}{(a) The \snet{} solution in \colorbox{train}{training} (\ie $g_{Sauvola}(\cdot)$) and \colorbox{test}{testing} (\ie $f_{\snet{}}(\cdot)$)} \\
         \includegraphics[trim=0 3cm 62cm 39cm, clip, width=0.45\linewidth]{pics/Structure2.png} & \includegraphics[trim=57cm 3cm 2cm 39cm, clip, width=0.45\linewidth]{pics/Structure2.png} \\
         (b) Multi-Window Sauvola & (c) Pixel-wise Window Attention\\ 
    \end{tabular}
    \caption{The overview of \snet{} solution and its trainable modules. $g_{\textrm{Sauvola}}$ and $g_{\textrm{SauvolaNet}}$ indicate the customized \sauvola{} layer and \snet{}, respectively; \texttt{Conv2D} and \texttt{AtrousConv2D} indicate the traditional atrous (w/ dilation rate 2) convolution layers, respectively; each \texttt{Conv2D}/\texttt{AtrousConv2D} are denoted of format \texttt{filters}@\texttt{ksize}$\times$\texttt{ksize} and followed by \texttt{InstanceNorm} and \texttt{ReLU} layers; the last \texttt{Conv2D} in window attention uses the \texttt{Softmax} activation (denoted w/ \colorbox{yellow}{borders}); and Pixelwise Thresholding indicates the binarization process \eqref{eq:bij}. 
    }
    \label{fig:structure}
\end{figure}

Fig.~\ref{fig:structure} describes the proposed \snet{} solution. It learns an auxiliary threshold estimation function from data by using a dual-branch design with three main modules, namely Multi-Window Sauvola (MWS), Pixelwise Window Attention (PWA), and Adaptive Sauvola Threshold (AST).

Specifically, the MWS module takes a gray-scale input image $\mathbf{D}$ and leverages on the \sauvola{} to compute the local thresholds for different window sizes. The PSA module also takes $\mathbf{D}$ as the input but estimates the attentive window sizes for each pixel location. The AST module predicts the final threshold for each pixel location $\mathbf{T}$ by fusing the thresholds of different window sizes from MWS using the attentive weights from PWA. As a result, the proposed \snet{} is analogous to a multi-window \sauvola{}, and models an auxiliary threshold estimation function $g_{\texttt{SauvolaNet}}$ between the input $\mathbf{D}$ and the output $\mathbf{T}$ as follows,
\begin{equation}\label{eq:g_sauvolanet}
    \mathbf{T}=g_{\texttt{SauvolaNet}}(\mathbf{D})
\end{equation}
Unlike in the classic \sauvola{}'s threshold estimation function \eqref{eq:g_sauvola}, \snet{} is end-to-end trainable and doesn't require any hyper-parameter. Similar to Eq.~\eqref{eq:fg}, the binarization decision function $f_{\texttt{SauvolaNet}}$ used in testing as shown below 
\begin{equation}
    \hat{\mathbf{B}} = f_{\texttt{SauvolaNet}}(\mathbf{D}) = th(\mathbf{D}, \mathbf{T}) = th(\mathbf{D}, g_{\texttt{SauvolaNet}}(\mathbf{D}))
\end{equation}
and the extra thresholding process (\ie \eqref{eq:bij}) is denoted as the Pixelwise Thresholding (PT) in Fig.~\ref{fig:structure}. Details about these modules are discussed in later sections.

\begin{figure}[!t]
    \centering\scriptsize
    \def\lw{0.26}
    \newcommand{\qedsymbol}{\rule{0.75em}{0.75em}}
    \begin{tabular}{c|l|c}\hline
    & \textbf{Single-window Threshold} & \textbf{Binarized Result} \\\hline
     \includegraphics[width=\lw\linewidth]{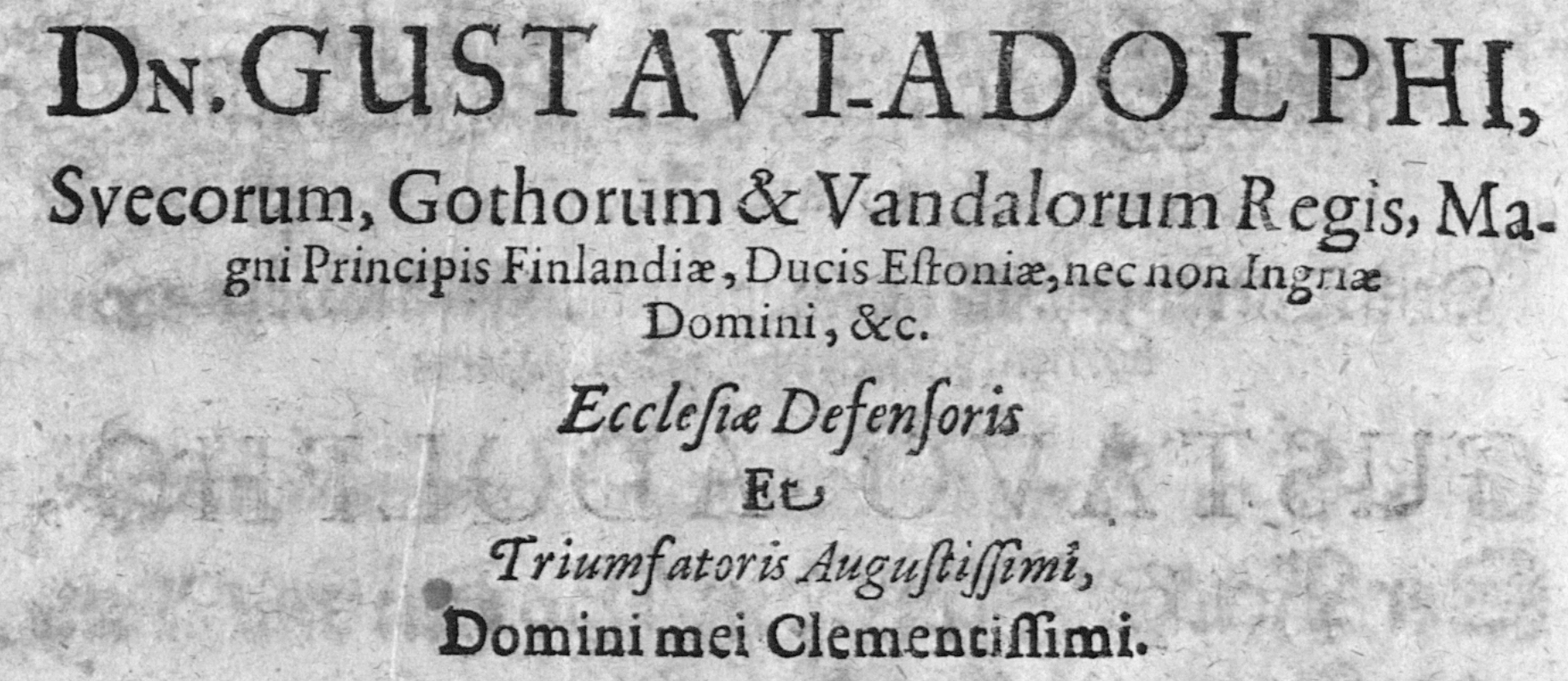}
         & \includegraphics[width=\lw\linewidth]{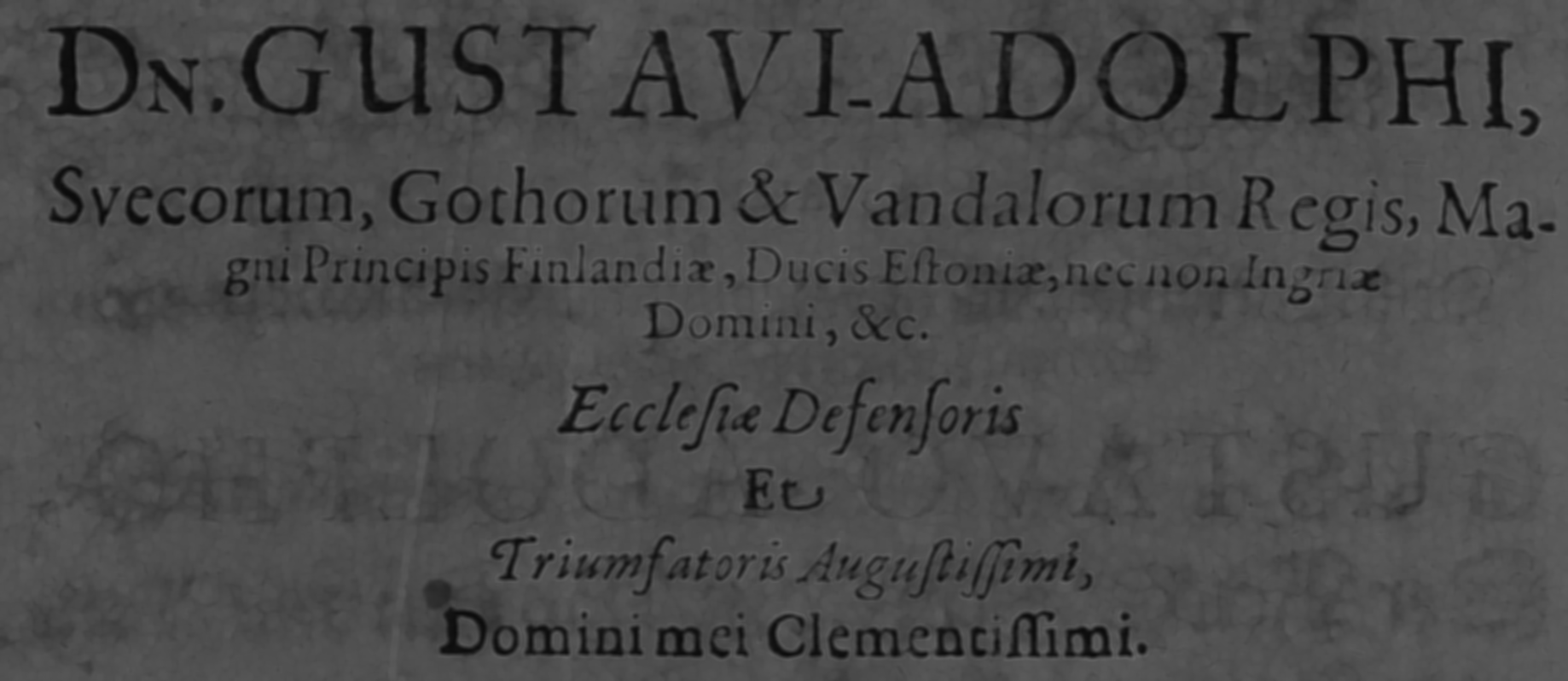}& \includegraphics[width=\lw\linewidth]{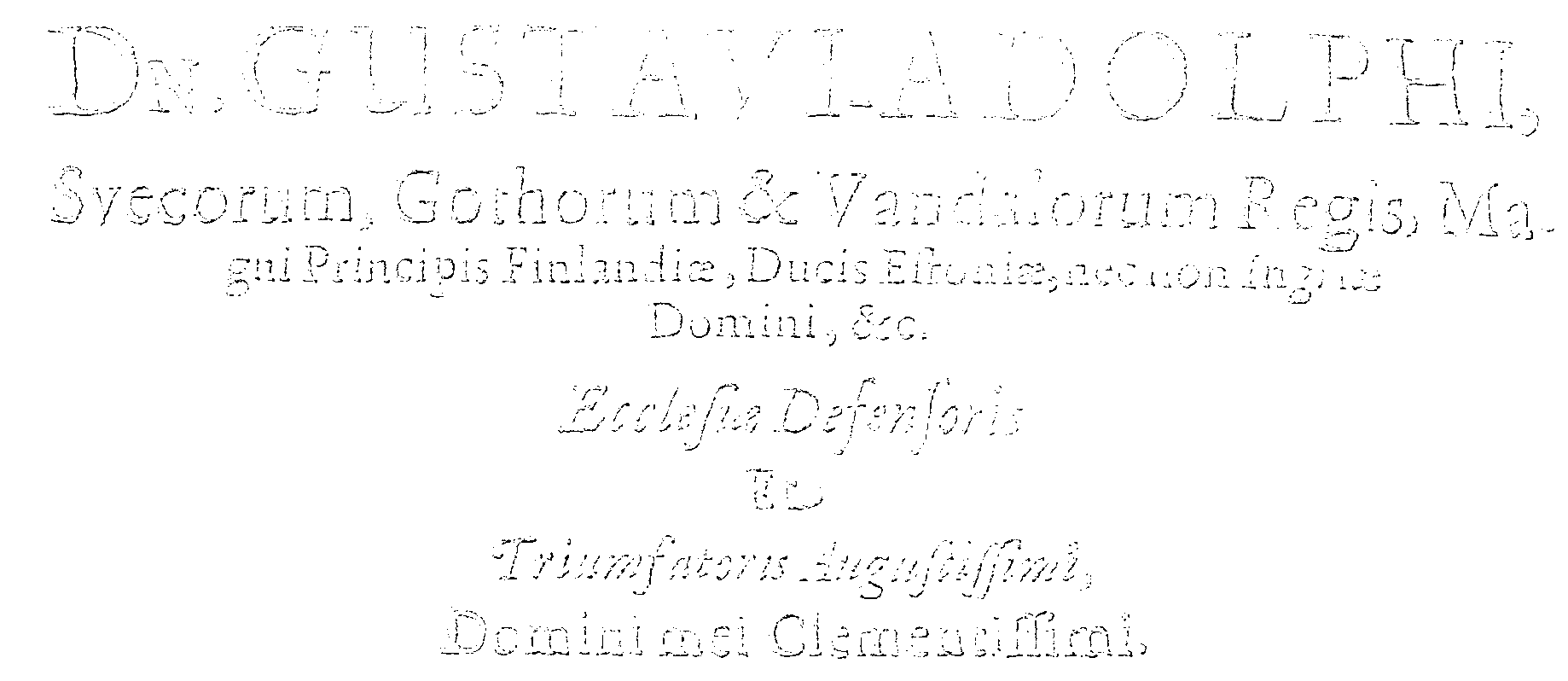}   \\
    (a) Input $\mathbf{D}$ 
         & {\textcolor[rgb]{0,0,0.5}{\qedsymbol}} (b1) $\mathbf{S}_1=g_{\texttt{Sauvola}|{w=7}}(\mathbf{D})$ & (c1) $th(\mathbf{D},\mathbf{S}_1)$ \\\cline{1-1}
         & \includegraphics[width=\lw\linewidth]{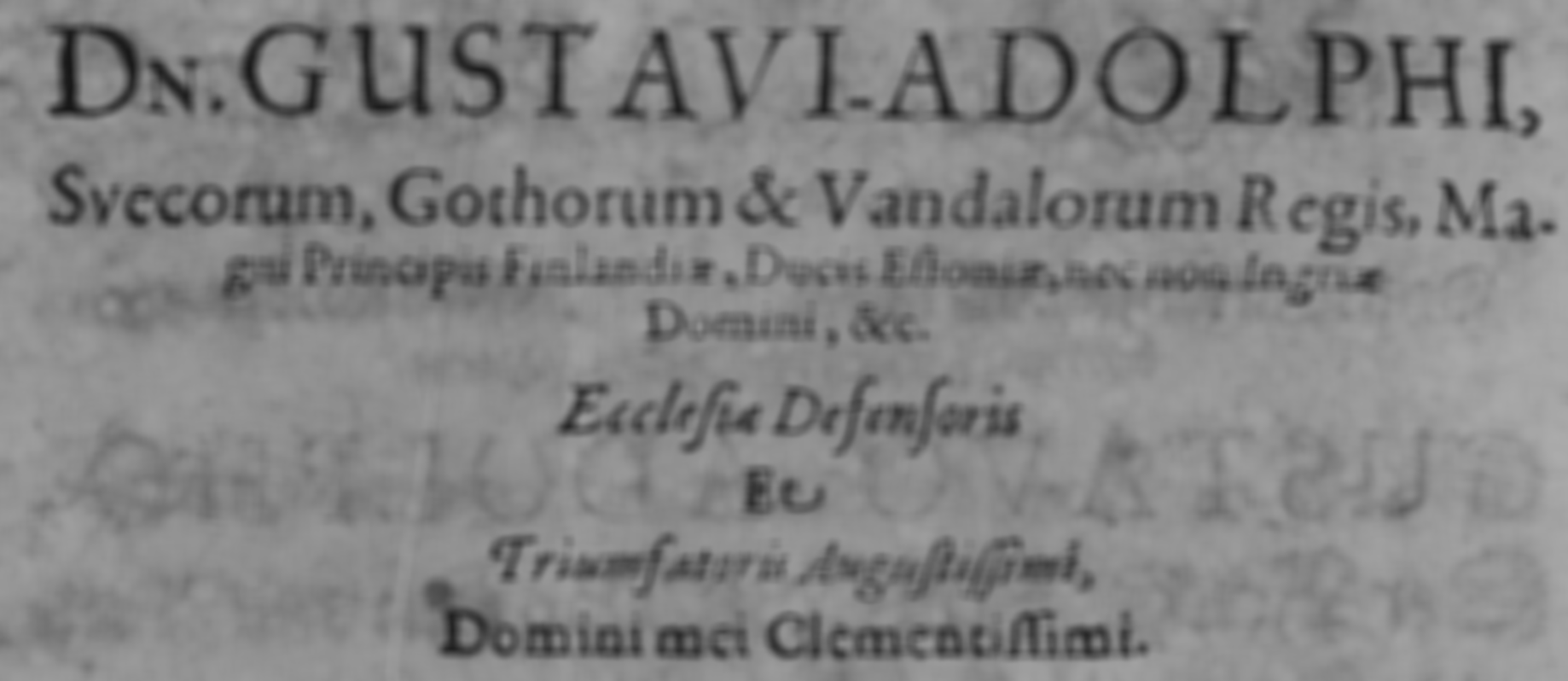}& \includegraphics[width=\lw\linewidth]{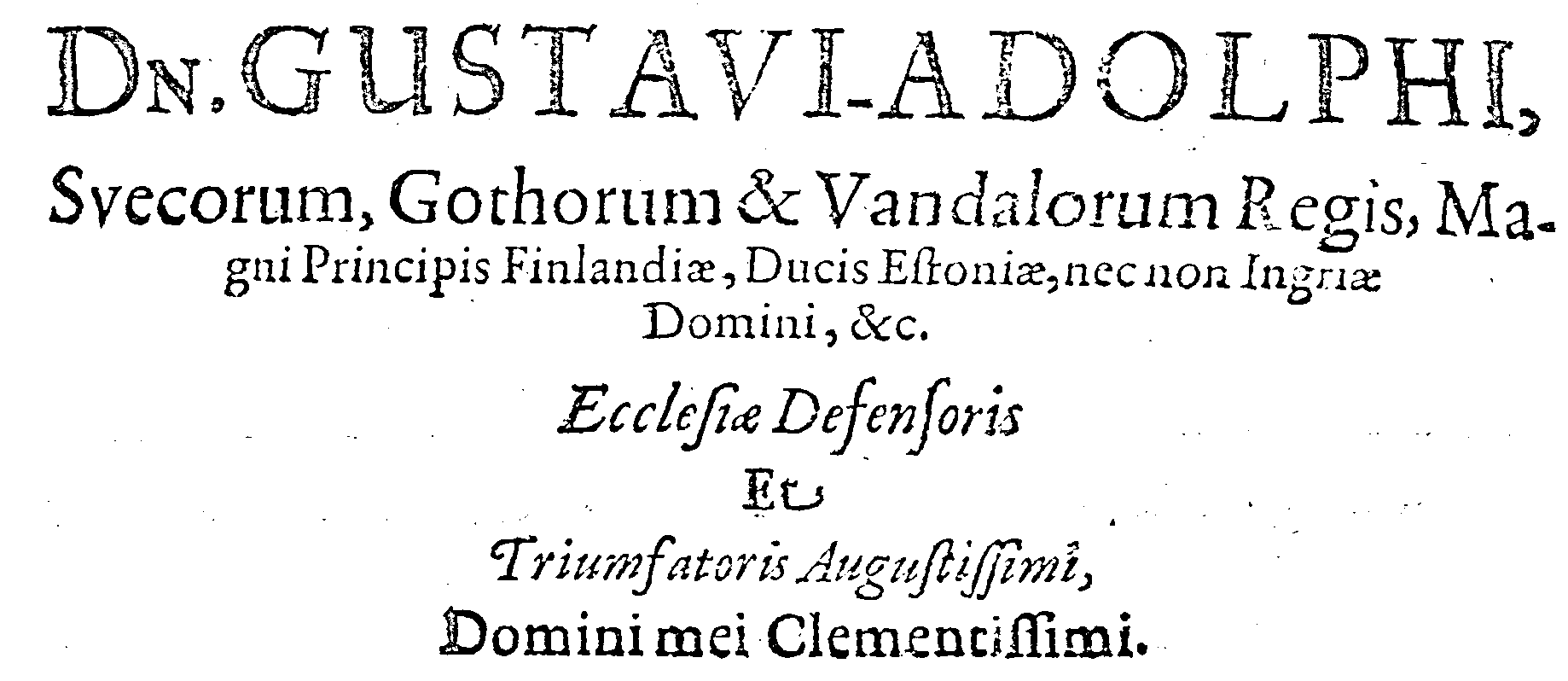} \\
         & {\textcolor[rgb]{0,0.0647,1.0}{\qedsymbol}} (b2) $\mathbf{S}_2=g_{\texttt{Sauvola}|{w=15}}(\mathbf{D})$ & (c2) $th(\mathbf{D},\mathbf{S}_2)$ \\
         & \includegraphics[width=\lw\linewidth]{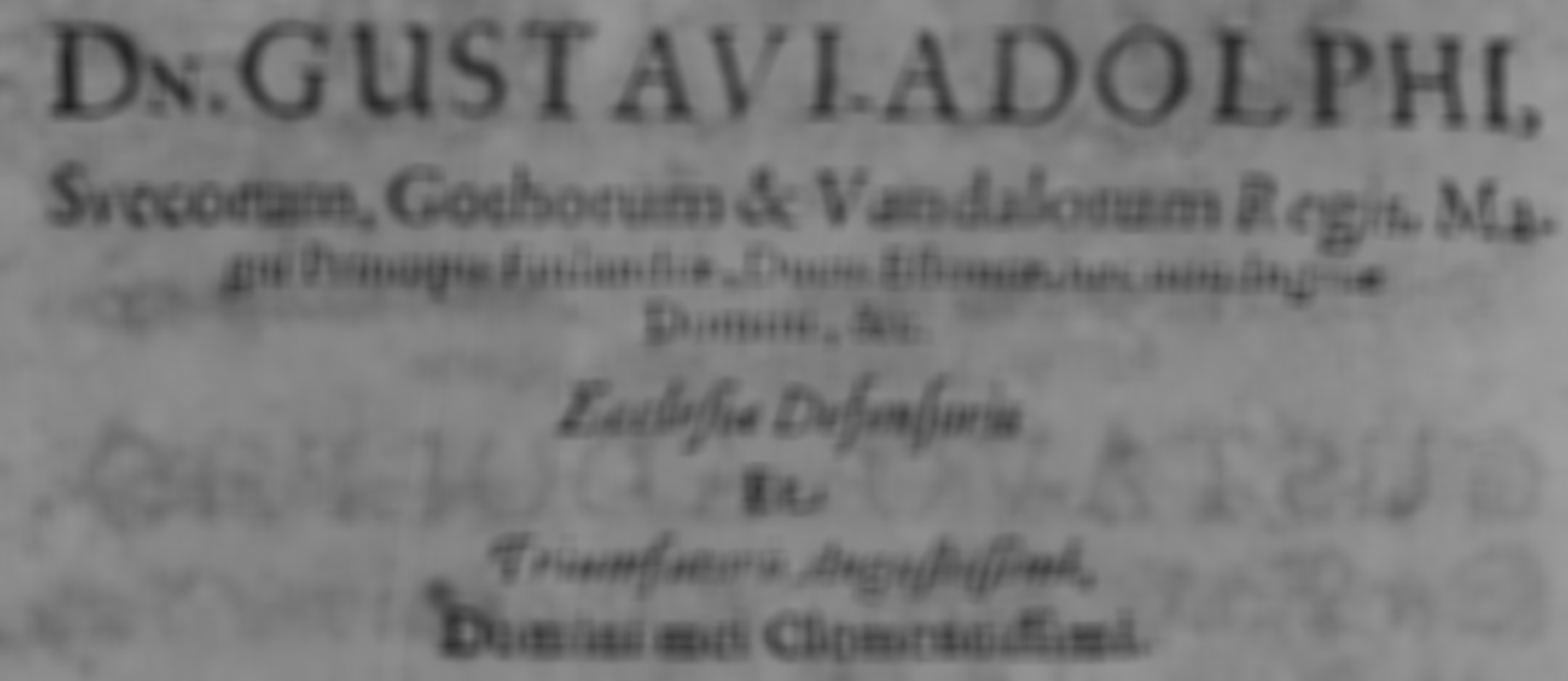}& \includegraphics[width=\lw\linewidth]{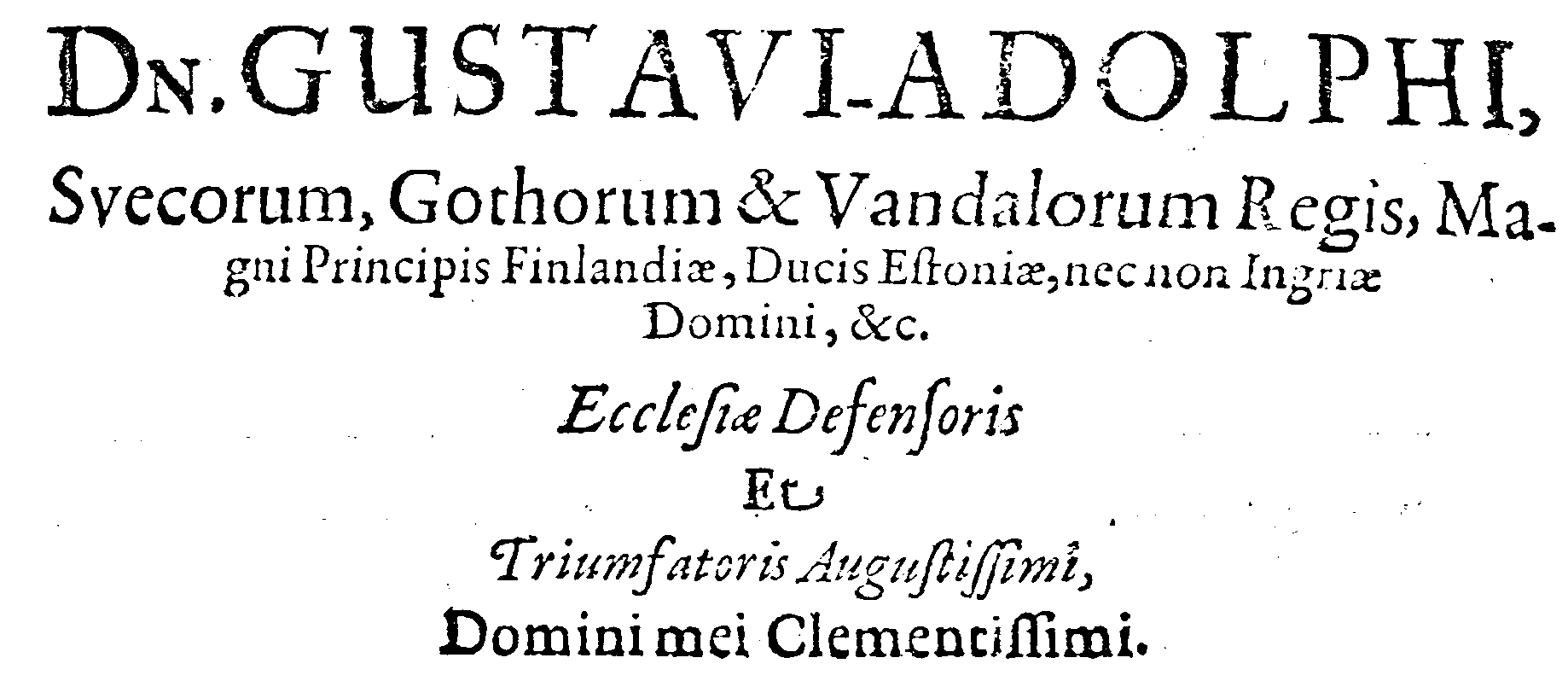} \\
         & {\textcolor[rgb]{0,0.645,1.0}{\qedsymbol}} (b3) $\mathbf{S}_3=g_{\texttt{Sauvola}|{w=23}}(\mathbf{D})$ & (c3) $th(\mathbf{D}, \mathbf{S}_3)$ \\
         & \includegraphics[width=\lw\linewidth]{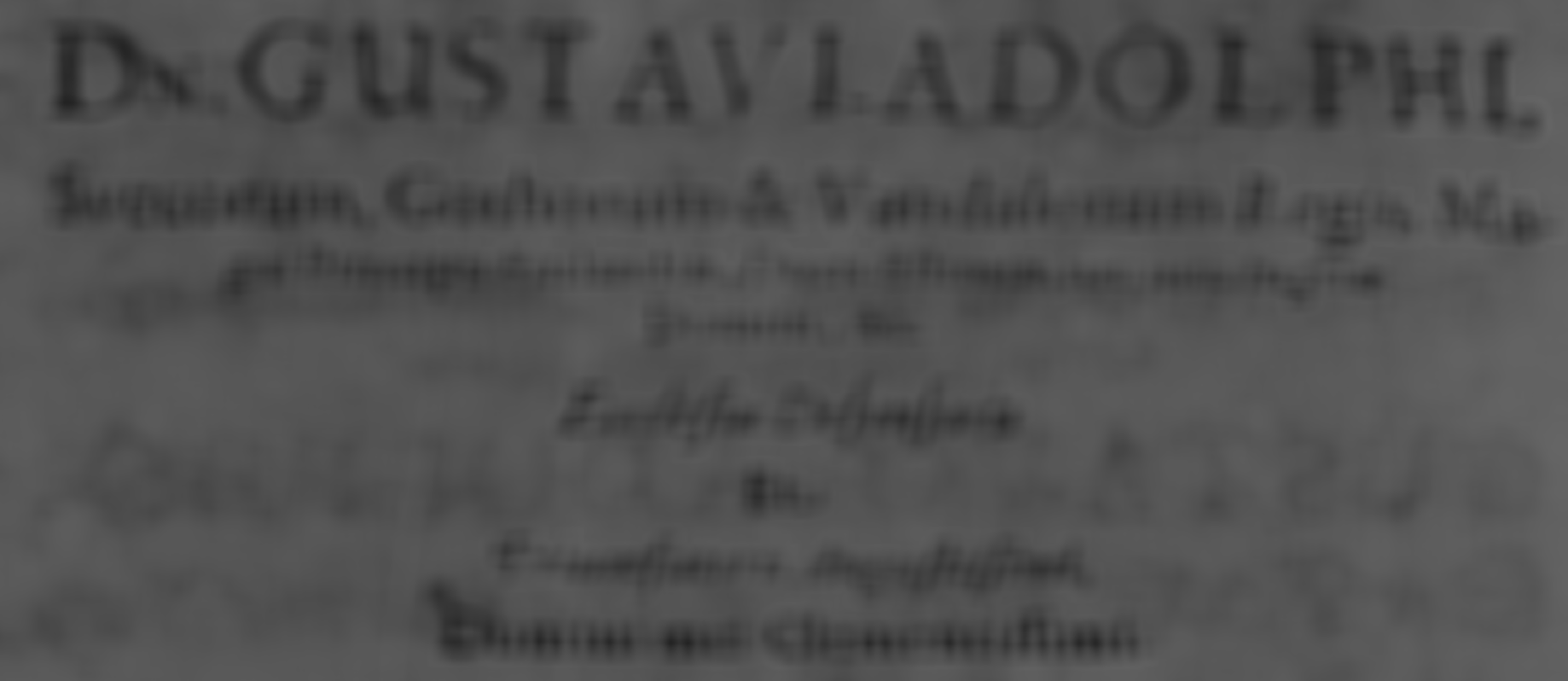}& \includegraphics[width=\lw\linewidth]{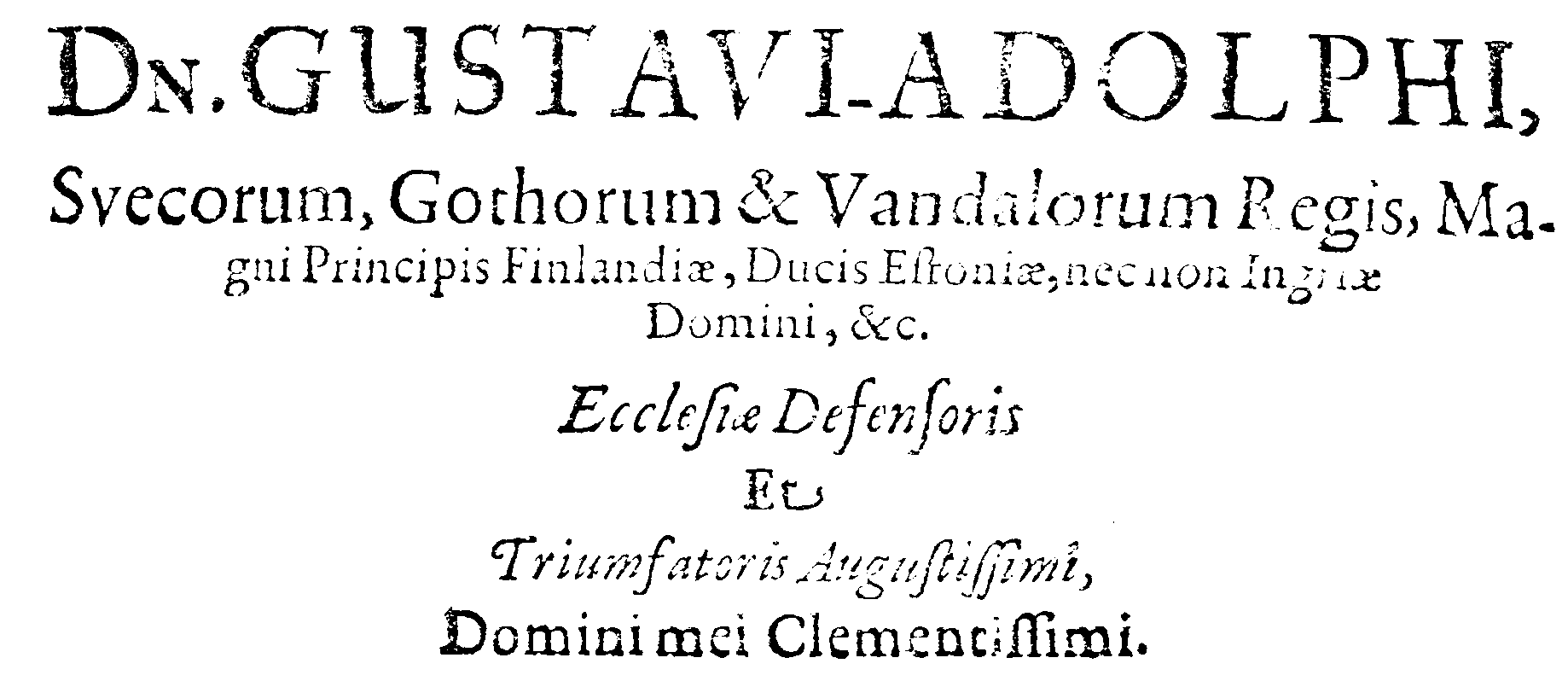} \\
         & {\textcolor[rgb]{0.2498,1,0.7179}{\qedsymbol}} (b4) $\mathbf{S}_4=g_{\texttt{Sauvola}|{w=31}}(\mathbf{D})$ & (c4) $th(\mathbf{D}, \mathbf{S}_4)$ \\
         & \includegraphics[width=\lw\linewidth]{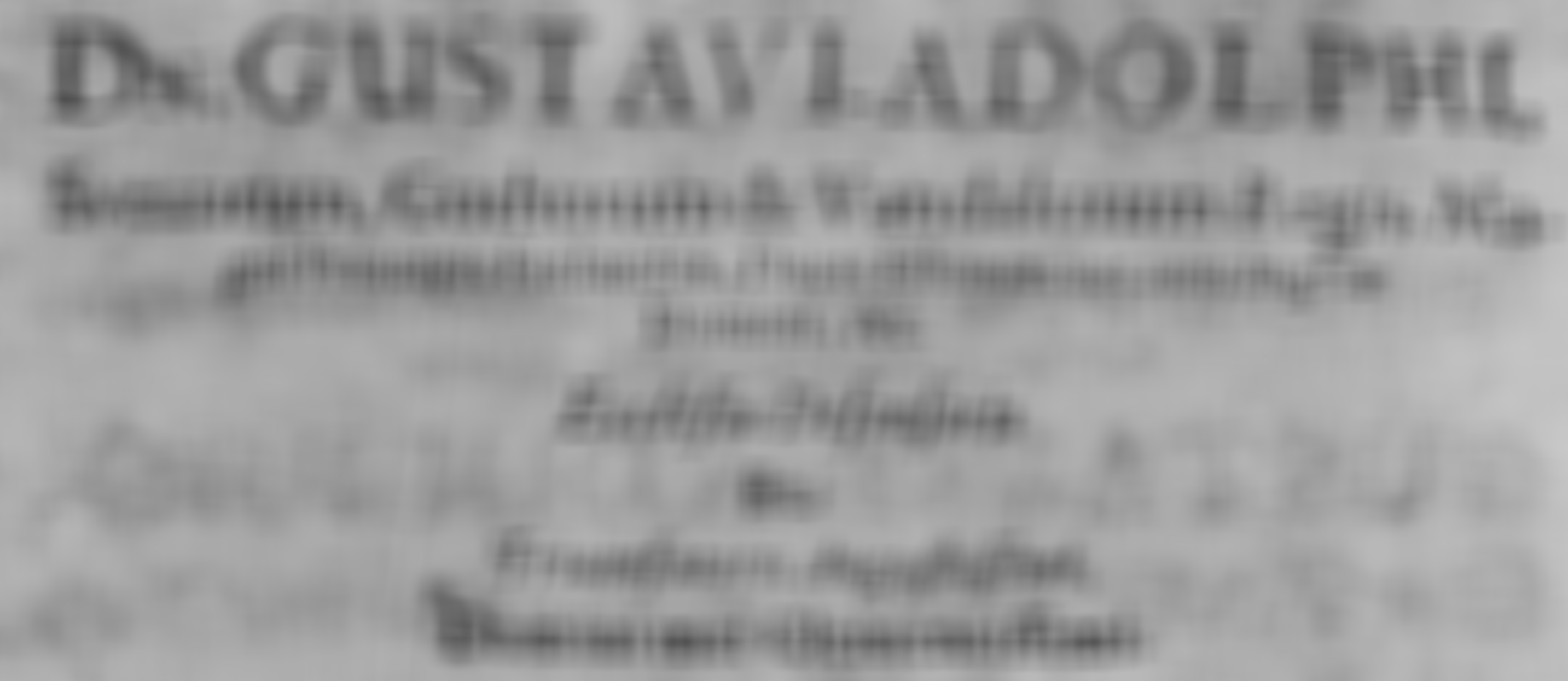}& \includegraphics[width=\lw\linewidth]{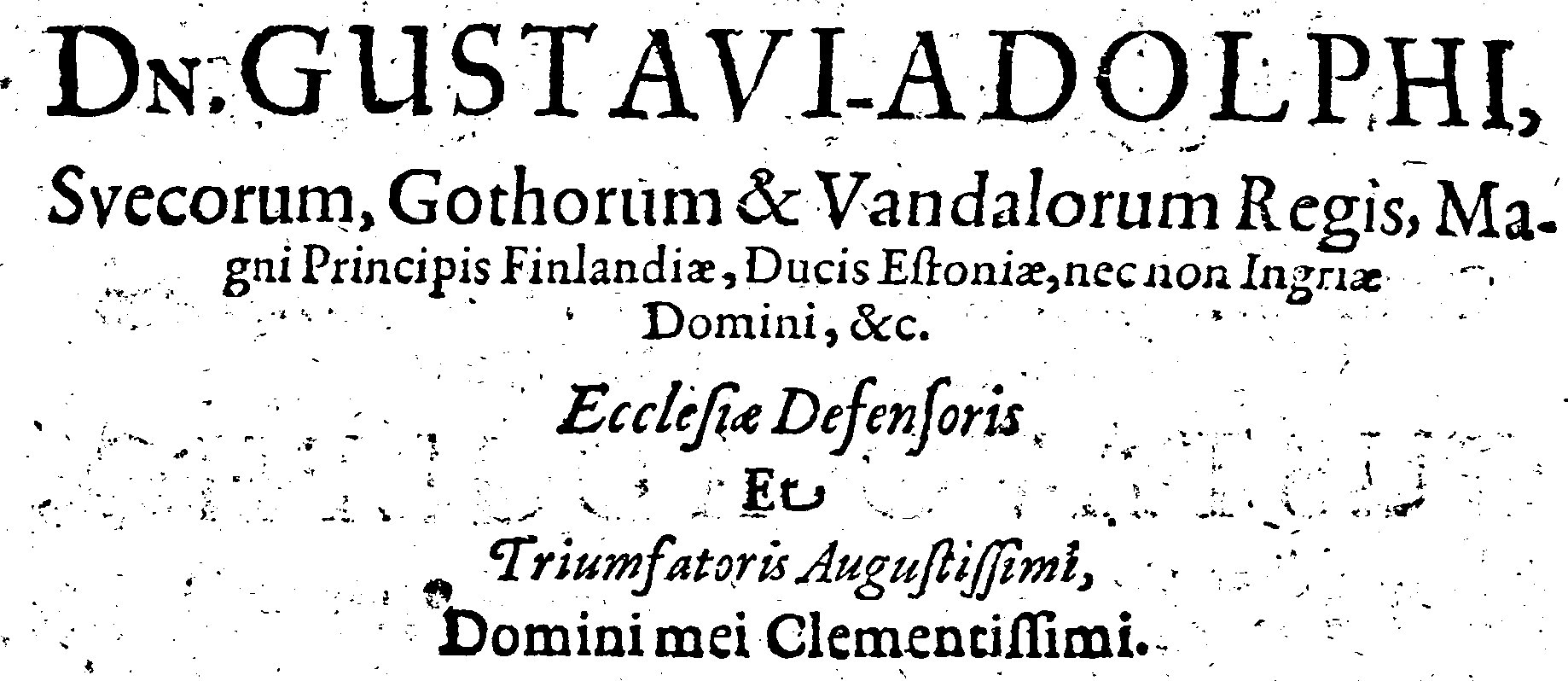} \\
         & {\textcolor[rgb]{0.7179,1.0, 0.2498}{\qedsymbol}} (b5) $\mathbf{S}_5=g_{\texttt{Sauvola}|{w=39}}(\mathbf{D})$ & (c5) $th(\mathbf{D}, \mathbf{S}_5)$ \\
         & \includegraphics[width=\lw\linewidth]{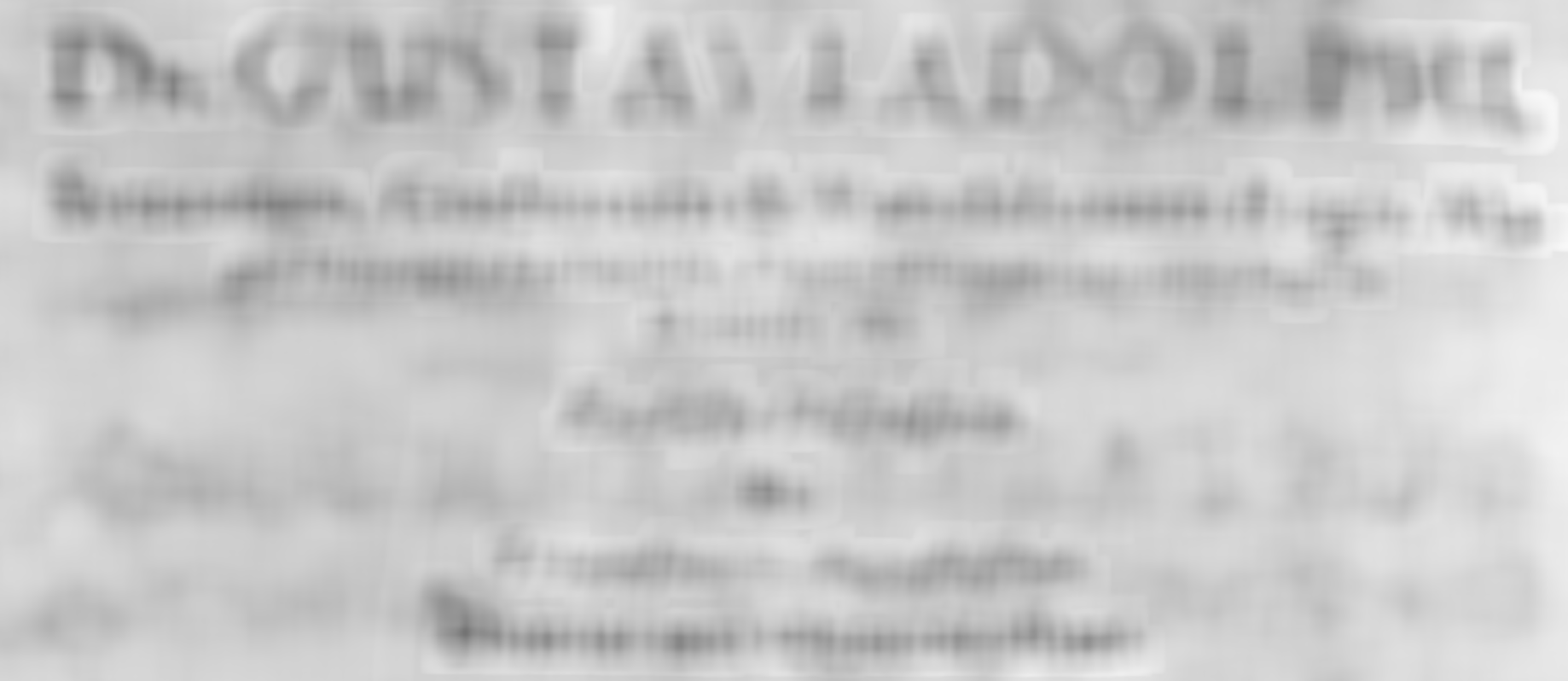}& \includegraphics[width=\lw\linewidth]{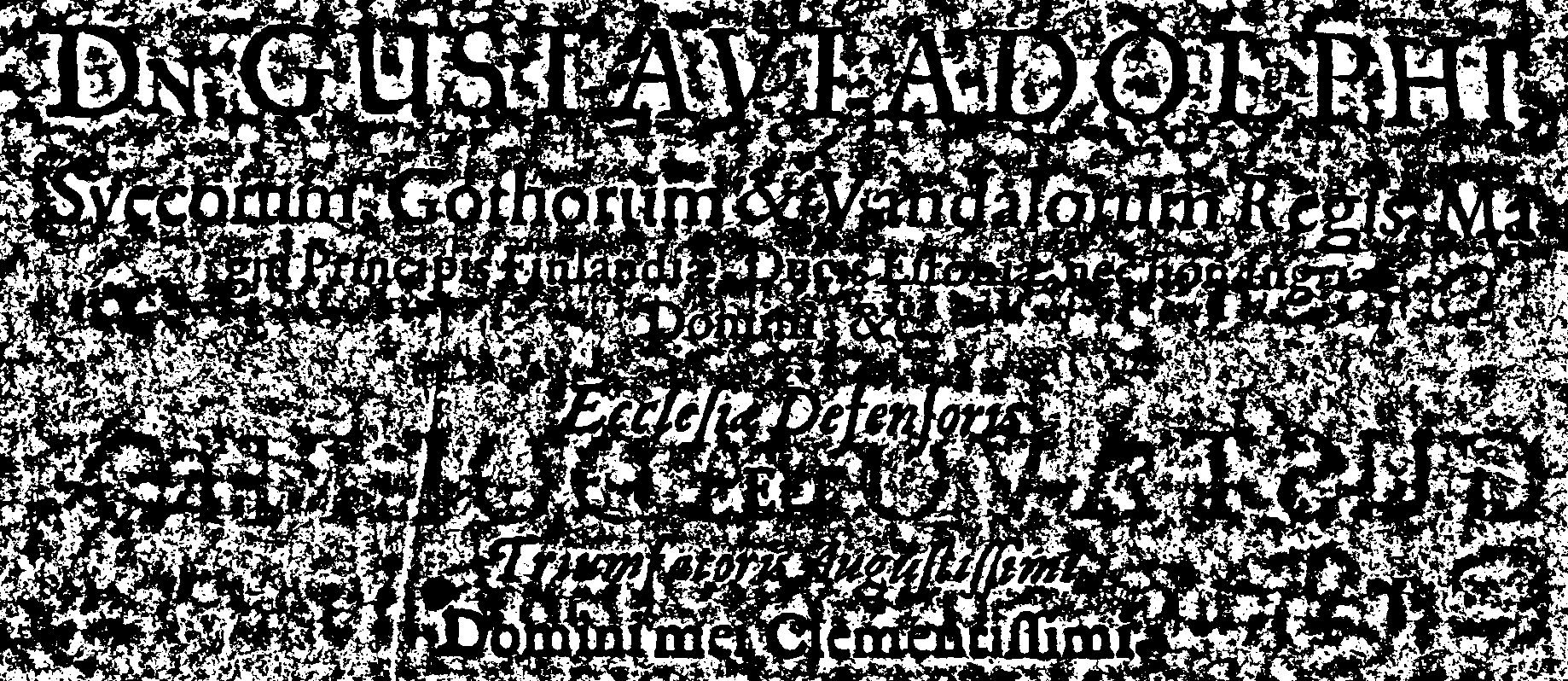} \\
         & {\textcolor[rgb]{1,0.7269,0.}{\qedsymbol}} (b6) $\mathbf{S}_6=g_{\texttt{Sauvola}|{w=47}}(\mathbf{D})$ & (c6) $th(\mathbf{D}, \mathbf{S}_6)$ \\\cline{1-1}
         \includegraphics[width=\lw\linewidth]{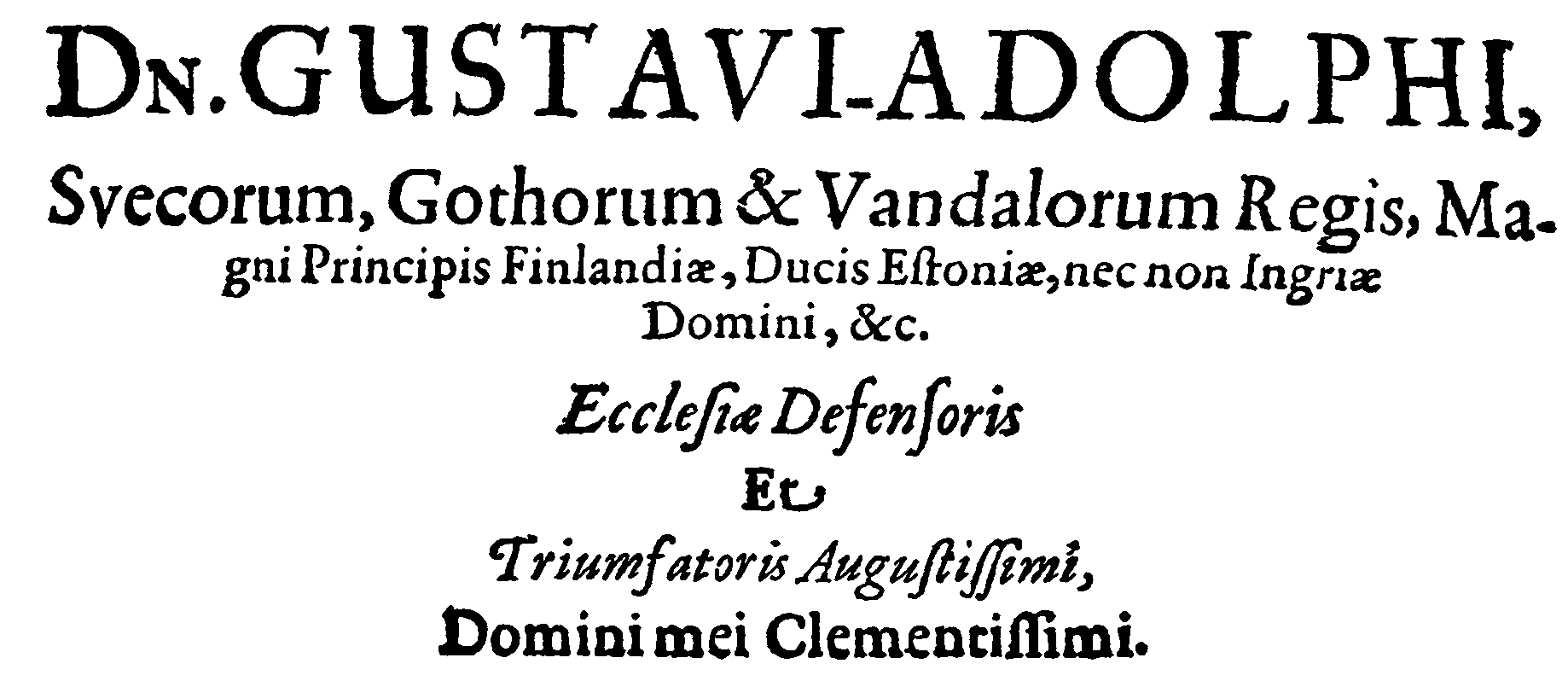}
         & \includegraphics[width=\lw\linewidth]{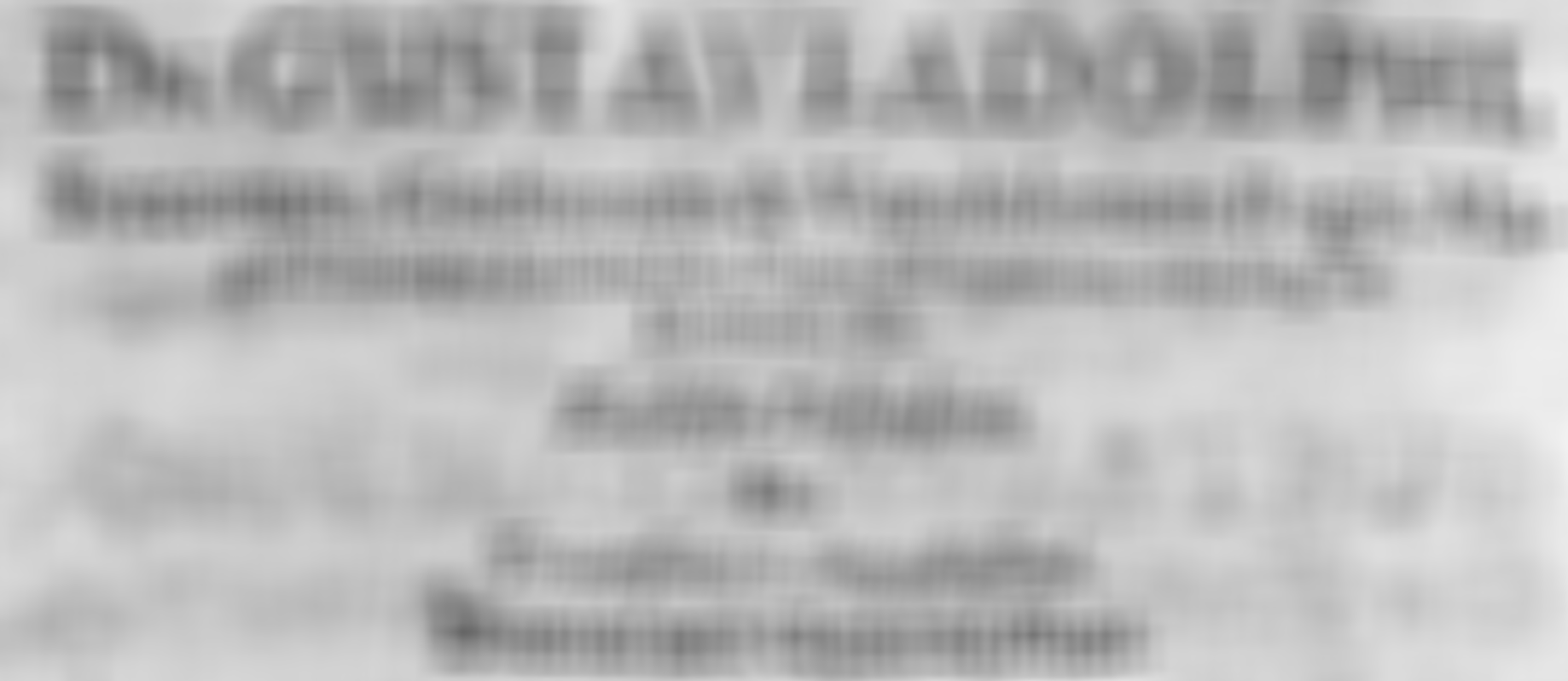}& \includegraphics[width=\lw\linewidth]{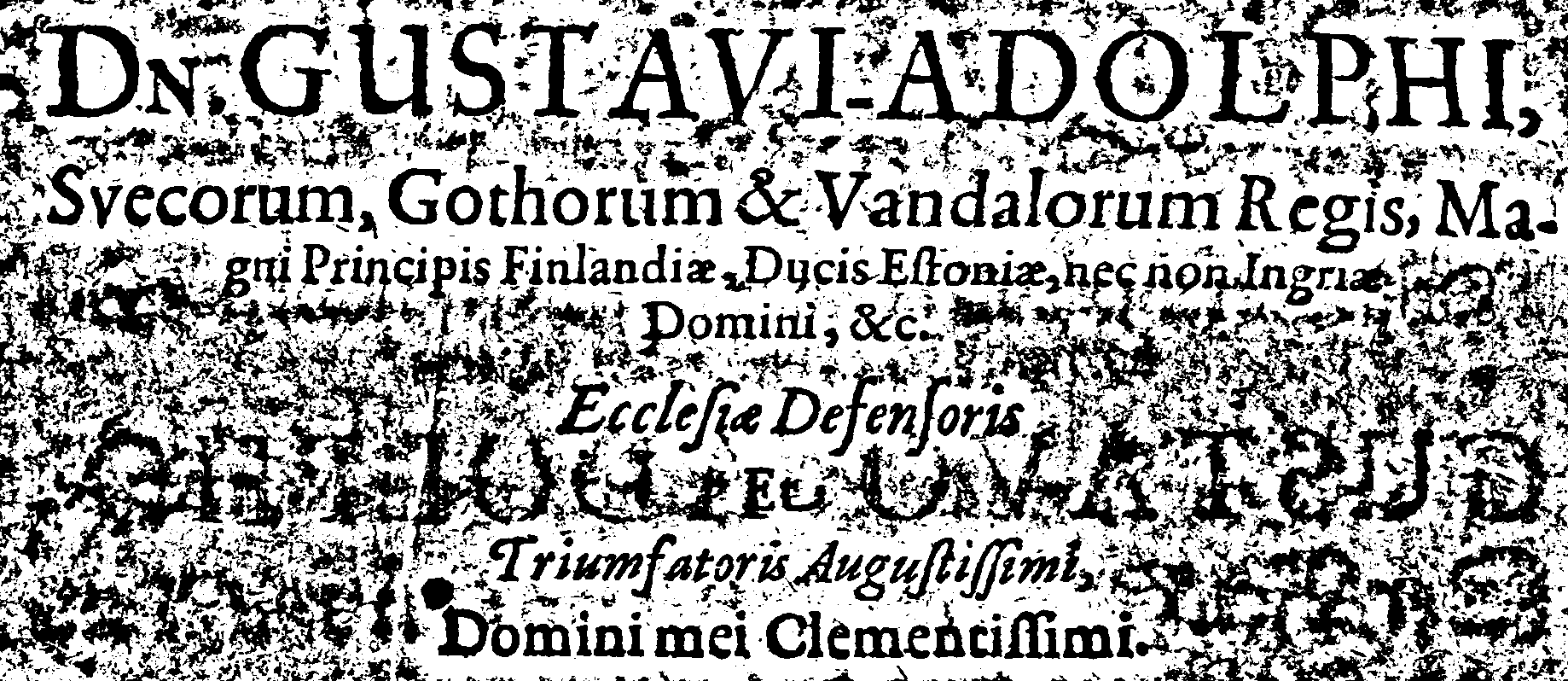} \\
         (d) GroundTruth & {\textcolor[rgb]{1,0.1895,0.}{\qedsymbol}} (b7) $\mathbf{S}_7=g_{\texttt{Sauvola}|{w=55}}(\mathbf{D})$ & (c7) $th(\mathbf{D}, \mathbf{S}_7)$ \\\cline{1-1}
    \includegraphics[width=\lw\linewidth]{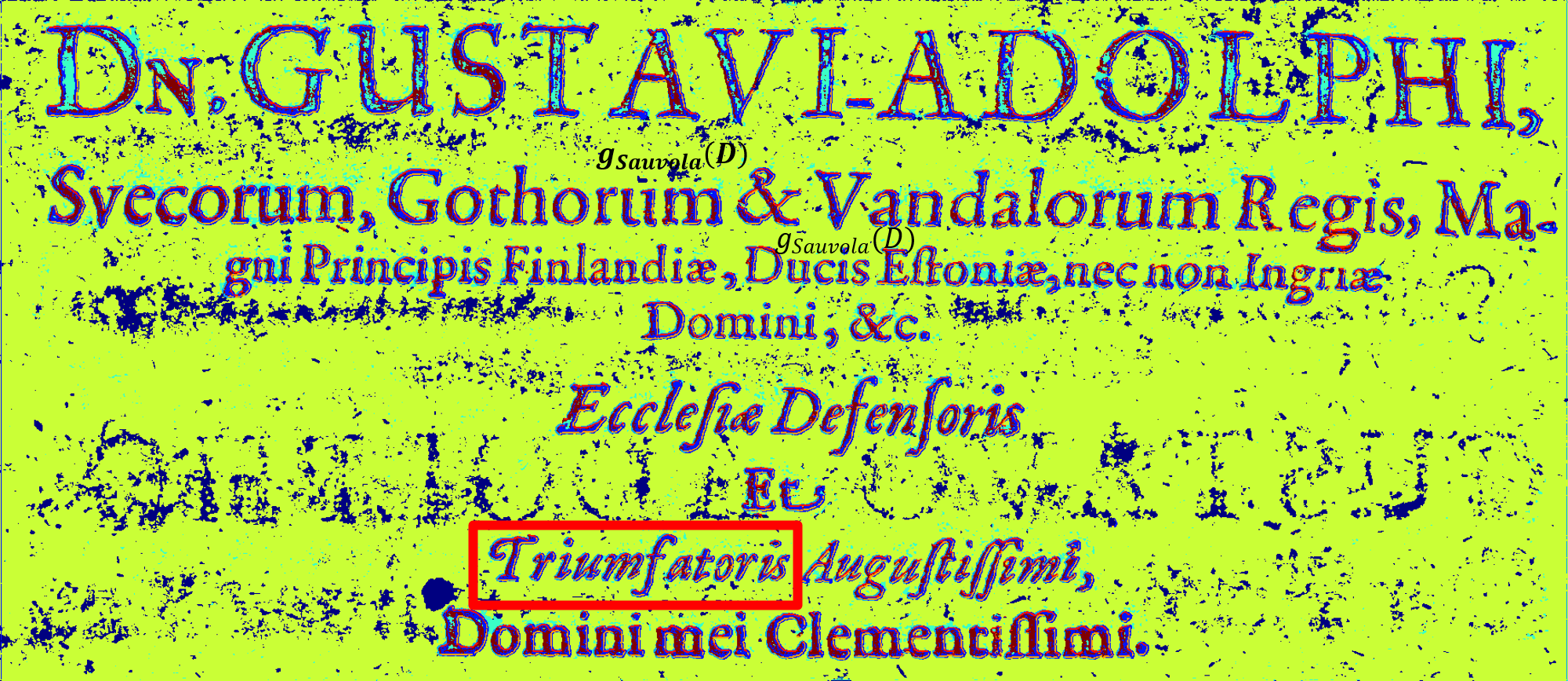} 
         & \includegraphics[width=\lw\linewidth]{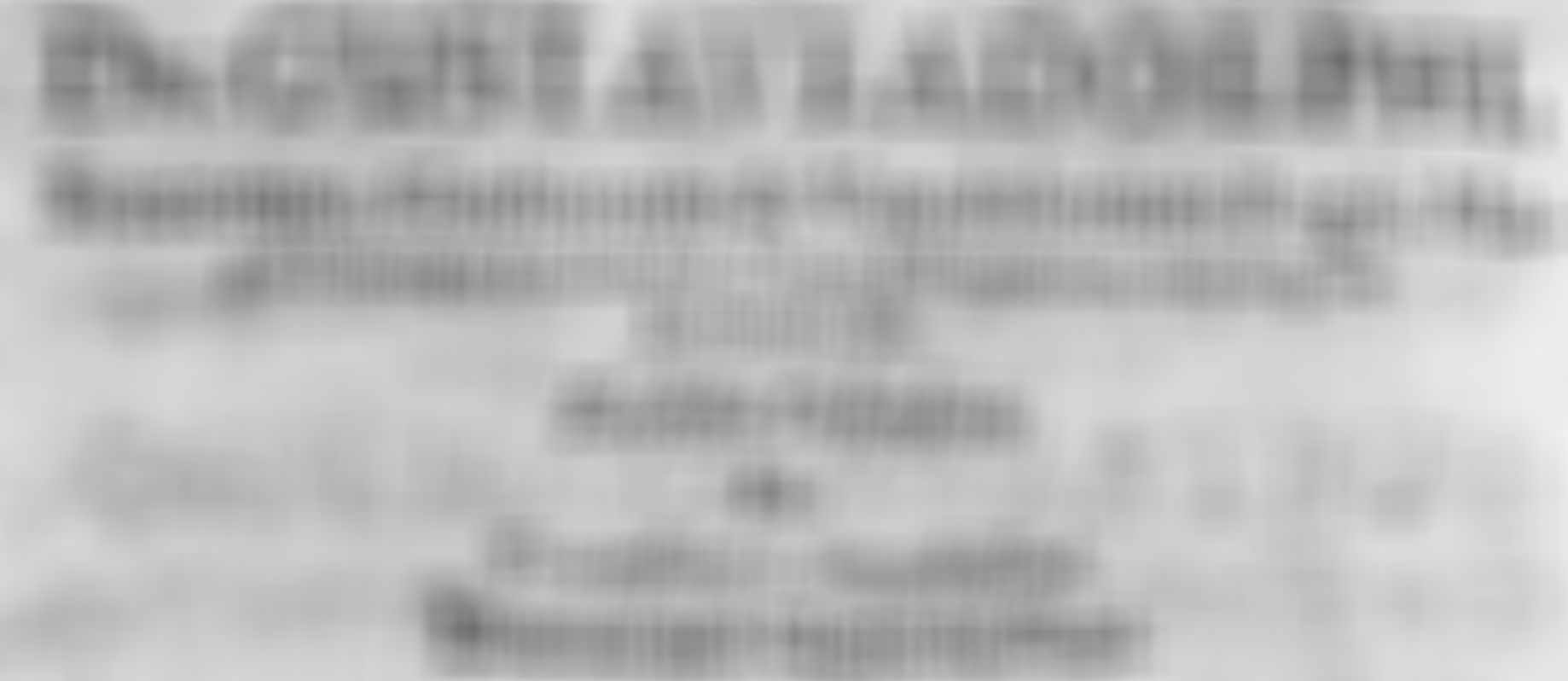}& \includegraphics[width=\lw\linewidth]{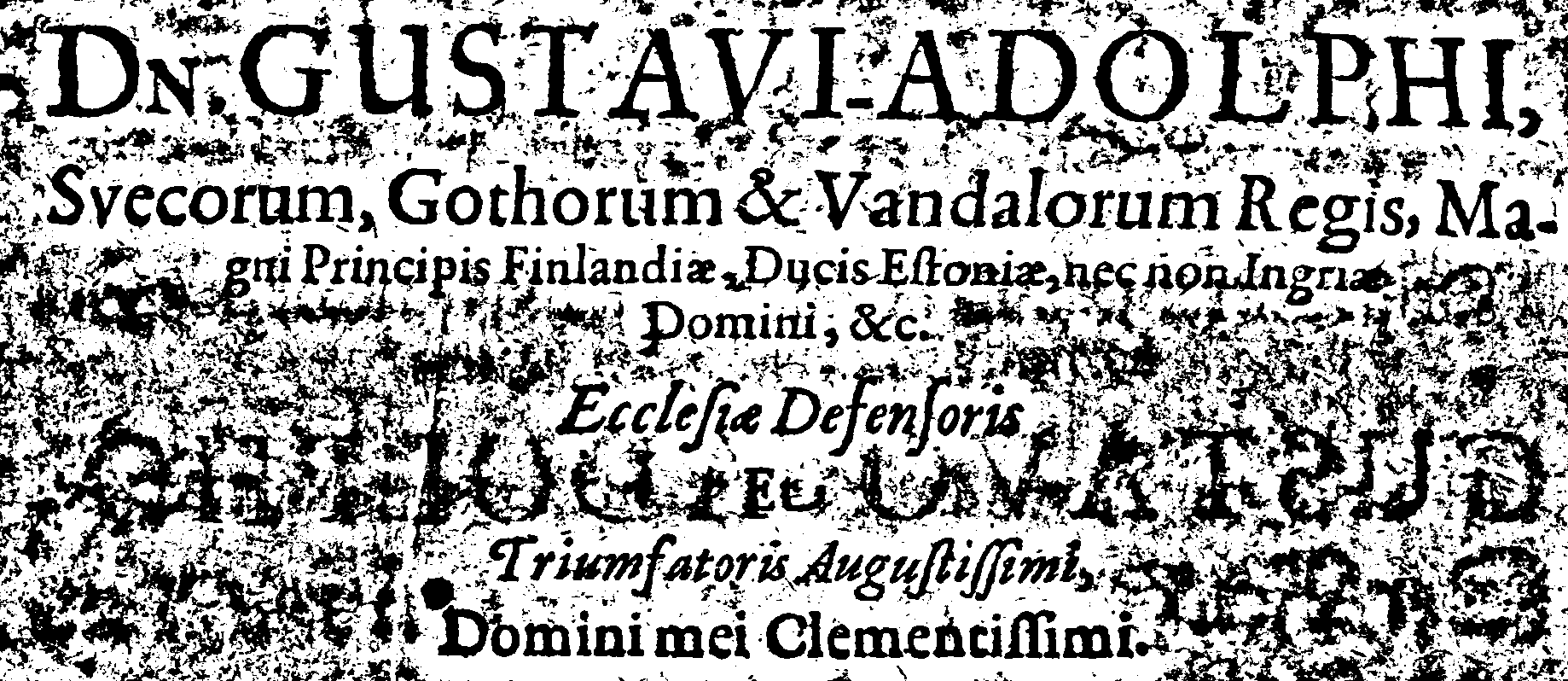} \\
    (e) Window attention  & {\textcolor[rgb]{0.5,0,0}{\qedsymbol}} (b8) $\mathbf{S}_8=g_{\texttt{Sauvola}|{w=63}}(\mathbf{D})$ & (c8) $th(\mathbf{D},\mathbf{S}_8)$ \\
    \hline
    \includegraphics[width=\lw\linewidth,height=1.35cm,trim=550 100 900 620,clip]{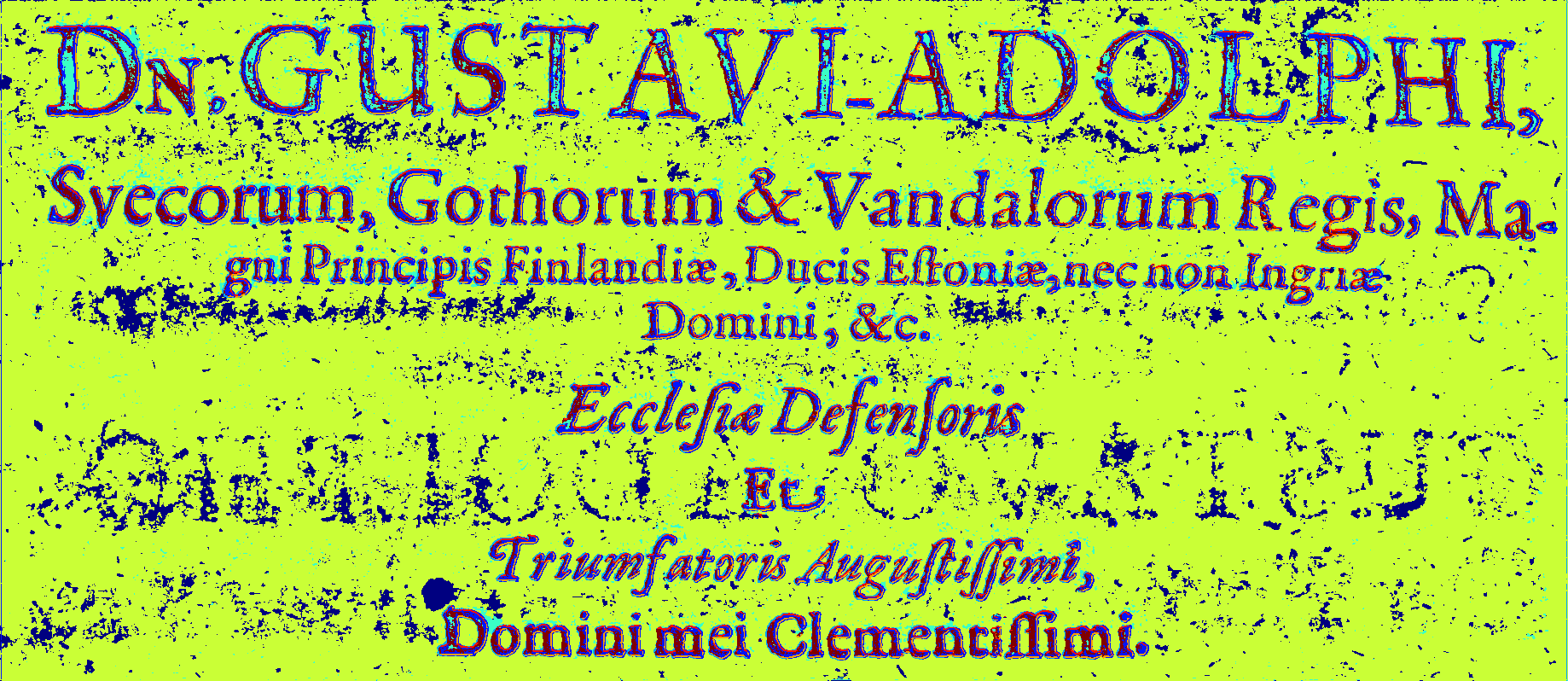} 
        & \includegraphics[width=\lw\linewidth]{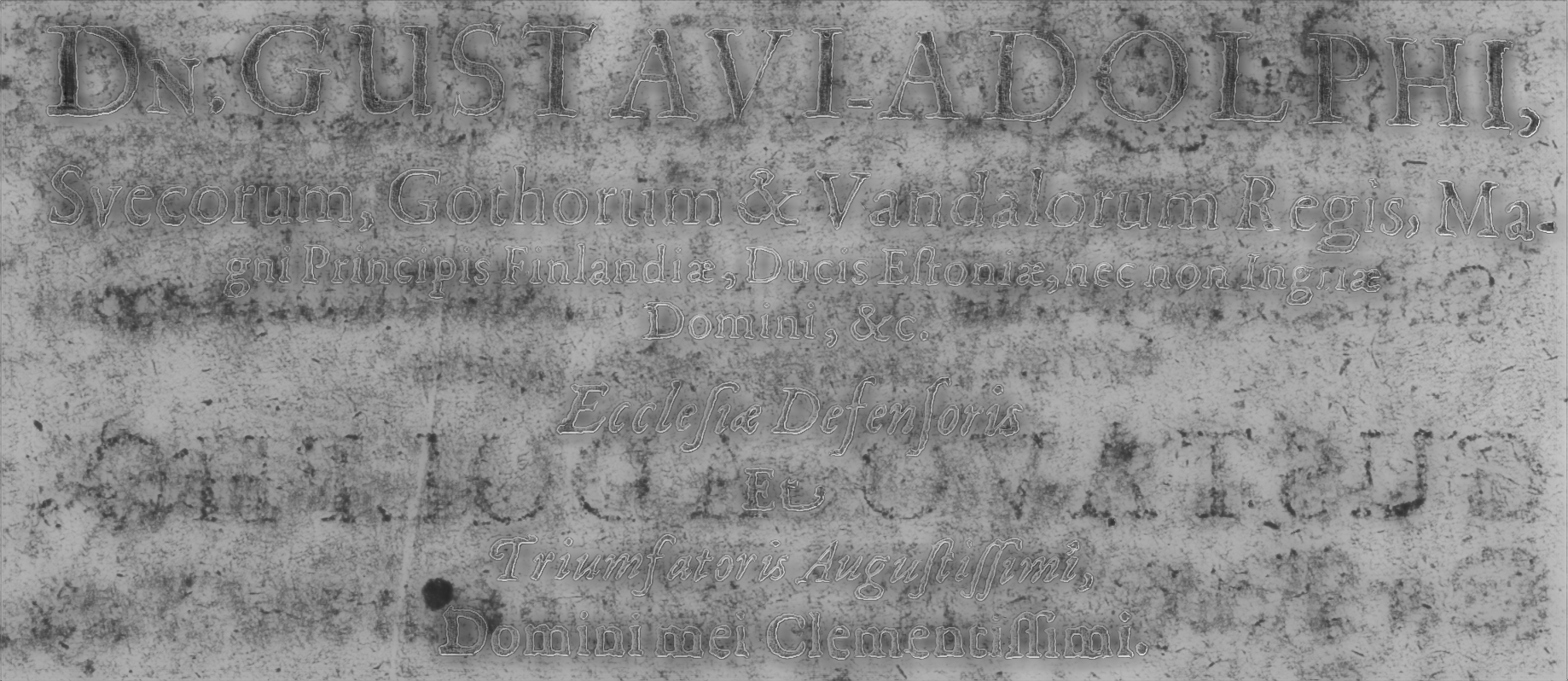} & \includegraphics[width=\lw\linewidth]{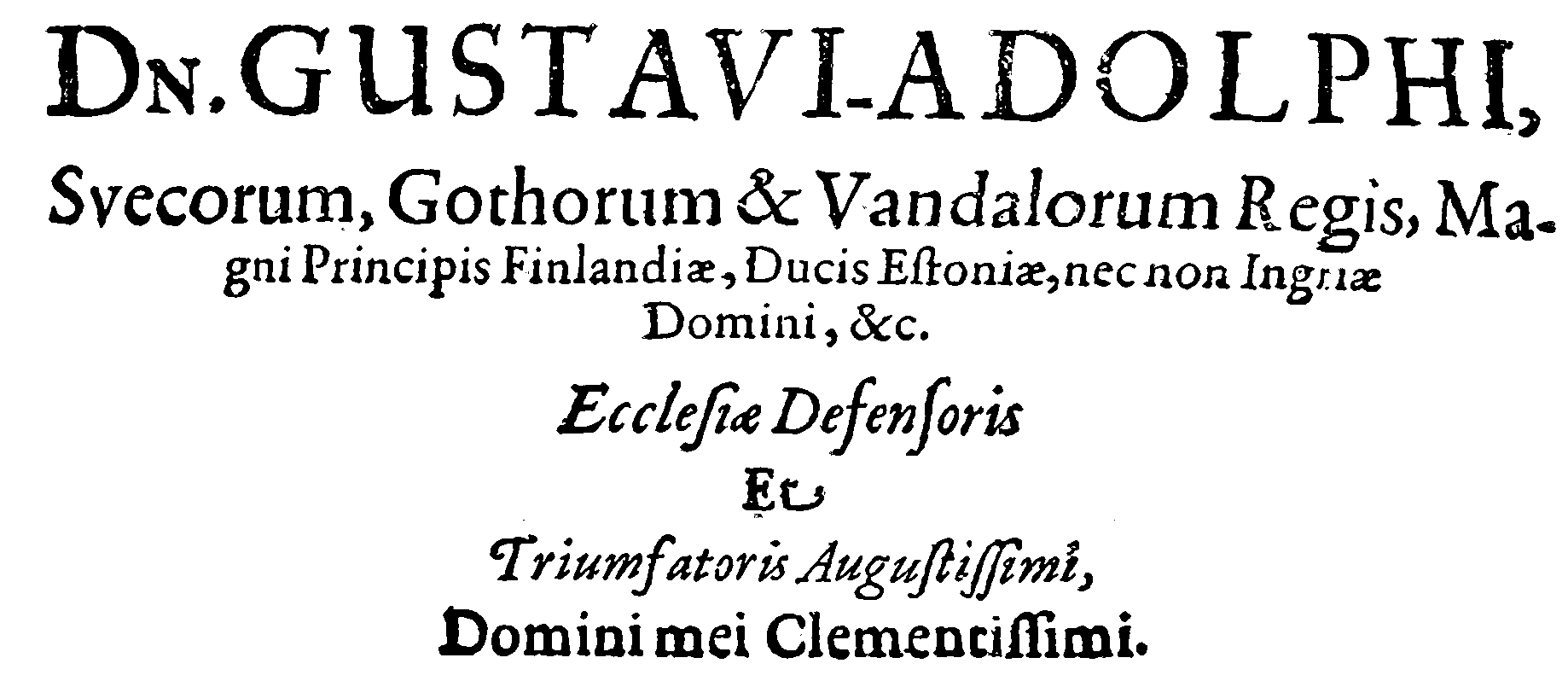} \\
    (f) 5x enlarged red zone in (e)
        & (g) Adaptive Threshold $\mathbf{T}$ & (h) Output $\hat{\mathbf{B}}=th(\mathbf{D}, \mathbf{T})$ \\ \hline
    \end{tabular}
\caption{Intermediate results of \snet{}. Please note: 1) window attention (e) visualizes the most preferred window size of $\mathbf{A}$ for each pixel locations (\ie {\small{$\argmax(\mathbf{A},\textrm{axis}=-1)$}}), and the 8 used colors correspond to those put before (b*); 2) binarized images (c*) are not used in \snet{} but for visualization only; and 3) $g_{\textrm{Sauvola}}(\cdot)$ and $th(\cdot, \cdot)$ indicates the \texttt{Sauvola} layer and the pixelwise thresholding function~\eqref{eq:bij}, respectively. 
    }
    \label{fig:snet}
\end{figure}

\subsection{Multi-Window Sauvola}\label{sec:sauvolaLayer}
The MWS module can be considered as a re-implementation of the classic multi-window \sauvola{} analysis in the DNN context. More precisely, we first introduce a new DNN layer called \texttt{Sauvola} (denoted as $g_{\textrm{Sauvola}}(\cdot)$ in the function form), which has the \sauvola{} window size as input argument and \sauvola{}'s hyper-parameter $s$ and $r$ as trainable parameters. 
To enable multi-window analysis, we use a set of \texttt{Sauvola} layers, each corresponding to one window size in \eqref{eq:winsizes}.The selection of window are verified in Sec.~\ref{sec:ablation}.
\begin{equation}\label{eq:winsizes}
    \mathbb{W} = \left\{w \,|\, w \in [7, 15, 23, 31, 39, 47, 55, 63]\right\}
\end{equation}
Fig.~\ref{fig:snet} visualizes all intermediate outputs of \snet{}, and Fig.~\ref{fig:snet}-(b*) show predicted \sauvola{} thresholds based on these window sizes, and Fig.~\ref{fig:snet}-(c*) further binarize the input image using corresponding thresholds. These results again confirm that satisfactory binarization performance can be achieved by \sauvola{} when the appropriate window size is used.

It is worthy to emphasize that \texttt{Sauvola} threshold computing window-wise mean and the standard deviation (see \eqref{eq:sauvola}) is very time-consuming when using the traditional DNN layers (\eg, \texttt{AveragePooling2D}), especially for a big window size (\eg, 31 or above). Fortunately, we implement our \texttt{Sauvola} layer by using integral image solution~\cite{shafait2008efficient} to reduce the computational complexity to $O(1)$. 


\subsection{Pixelwise Window Attention}
As mentioned in many works~\cite{lazzara2014efficient,kaur2020modified}, one disadvantage when using \sauvola{} algorithm is the tuning of hyper-parameters. Among all three hyperparameters, namely, the window size $w$, the degradation level $k$, and the input deviation $r$, $w$ is the most important. Existing works typically decompose an input image into non-overlapping regions~\cite{pai2010adaptive,lazzara2014efficient} or grids~\cite{moghaddam2010multi} and apply each a different window size. However, existing solutions are not suitable for DNN implementation for two reasons: 1) non-overlapping decomposition is not a differentiable operation; and 2) processing regions/grids of different sizes are hard to parallelize. 

 Alternatively, we adopt the widely-accepted attention mechanism to remove the dependency on the user-specified window sizes. Specifically, the proposed PWA module is a sub-network that takes an input document image $\mathbf{D}$ and predicts the pixel-wise attention on all predefined window sizes. It conceptually follows the multi-grid method introduced by \texttt{DeepLabv3}~\cite{chen2017rethinking} while using a fixed dilation rate at 2. Also, we use the \texttt{InstanceNormalization} instead of the common \texttt{BatchNormalization} to mitigate the overfitting risk caused by a small training dataset. The detailed network architecture is shown in Fig.~\ref{fig:structure}. 

Sample result of PWA can be found in Fig.~\ref{fig:snet}-(e). As one can see, the proposed PWA successfully predicts different window sizes for different pixels. More precisely, it prefers $w=39$ (see Fig.~\ref{fig:snet}-(b5)) and $w=15$ (see Fig.~\ref{fig:snet}-(b2)) for background and foreground pixels, respectively; and uses very large window sizes, \eg, $w=63$ (\ie Fig.~\ref{fig:snet}-(b8)) for those pixels on text borders. 

\subsection{Adaptive Sauvola Threshold}
As one can see from Fig.~\ref{fig:structure}, the MWS outputs a Sauvola tensor $\mathbf{S}$ of size $H\times W\times N$, where $N$ is the number of used window sizes (and we use $N=8$, see Eq.~\eqref{eq:winsizes}), the PWA outputs an attention tensor $\mathbf{A}$ of the same size as $\mathbf{S}$ and the attention sum for all window sizes on each pixel location is always 1, namely, 
\begin{equation}\label{eq:sum1}
    \sum_{k=1}^{N}A[i,j,k] = 1, \,\, \forall 1\leq i\leq H,1\leq j\leq W. 
\end{equation}
The AST applies the window attention tensor $\mathbf{A}$ to the window-wise initial Sauvola threshold tensor $\mathbf{S}$ and compute the pixel-wise threshold $\mathbf{T}$ as below 
\begin{equation}\label{eq:thresh}
    T[i,j] = \sum_{k=1}^{N} A[i,j,k] \cdot S[i,j,k]
\end{equation}
Fig.~\ref{fig:snet}-(g) shows the adaptive threshold $\mathbf{T}$ when using the sample input Fig.~\ref{fig:snet}-(a). By comparing the corresponding binarized result (\ie Fig.~\ref{fig:snet}-(h)) with those of single window's results (\ie Fig.~\ref{fig:snet}-(c*)), one can easily verify that the adaptive threshold $\mathbf{T}$ outperforms any individual threshold result in $\mathbf{S}$.  


\subsection{Training, Inference, and Discussions}
In order to train \snet{}, we normalize the input $\mathbf{D}$ to the range of (0,1) (by dividing 255 for \texttt{uint8} image), and employ a modified hinge loss, namely

\begin{equation}\label{eq:loss}
    loss[i,j] = \max(1 - \alpha \cdot (D[i,j] - T[i,j])\cdot B[i,j], 0)
\end{equation}
where $\mathbf{B}$ is the corresponding binarization ground truth map with binary values -1 (foreground) and +1 (background); $\mathbf{T}$ is \snet{}'s predicted thresholds as shown in Eq.~\eqref{eq:g_sauvolanet}; and $\alpha$ is a parameter to empirically control the margin of decision boundary and only those pixels close to the decision boundary will be used in gradient back-propagation. Throughout the paper, we always use $\alpha=16$. 

We implement \snet{} in the \texttt{TensorFlow} framework. The used training patch size is 256$\times$256, and the data augmentations are random crop and random flip. The training batch size is set to 32, and we use \texttt{Adam} optimizer with the initial learning rate of 1$e$-3. During inference, we use $f_{\texttt{SauvolaNet}}$ instead of $g_{\texttt{SauvolaNet}}$ as shown in Fig.~\ref{fig:structure}-(a). It only differs from the training in terms of one extra the thresholding step \eqref{eq:bij} to compare \snet{} predicted thresholds with the original input to obtain the final binarized output. 

Unlike in most DNNs, each module in \snet{} is explainable: the MWS module leverages the \sauvola{} algorithm to reduce the number of required network parameters significantly, and the PWA module employs the attention idea to get rid of the \sauvola{}'s disadvantage of window size selection, and finally two branches are fused in the AST module to predict the pixel-wise threshold. Sample results in Fig.~\ref{fig:snet} further confirm that these modules work as expected.

\section{Experimental Results}\label{sec:experiment2}

\subsection{Dataset and Metrics}\label{sec:dataset}
In total, 13 document binarization datasets are used in experiments, and they are \{(H-)DIBCO 2009~\cite{gatos2009icdar} (10), 2010~\cite{pratikakis2010h} (10), 2011~\cite{Pratikakis2011icdar} (16), 2012~\cite{pratikakis2012icfhr} (14), 2013~\cite{pratikakis2013icdar} (16), 2014~\cite{ntirogiannis2014icfhr2014} (10), 2016~\cite{pratikakis2016icfhr2016} (10),  2017~\cite{pratikakis2017icdar2017} (20), 2018~\cite{Pratikakis2018icdar} (10); PHIDB~\cite{nafchi2013efficient} (15), Bickely-diary dataset~\cite{deng2010binarizationshop} (7), Synchromedia Multispectral dataset~\cite{hedjam2015icdar} (31), and Monk Cuper Set~\cite{he2019deepotsu} (25)\}. The braced numbers after each dataset indicates its sample size, and detailed partitions for training and testing will be specified in each study. For evaluation, we adopt the DIBCO metrics~\cite{Pratikakis2011icdar,pratikakis2012icfhr,pratikakis2013icdar,ntirogiannis2014icfhr2014,pratikakis2016icfhr2016,pratikakis2017icdar2017,Pratikakis2018icdar} namely, F-Measure (FM), psedudo F-Measure ($F_{ps}$), Distance Reciprocal Distortion metric (DRD) and Peak Signal-to-Noise Ratio (PSNR).

\subsection{Ablation studies}\label{sec:ablation}
To simplify discussion, let $\theta$ be the set of parameter settings related to a studied \sauvola{} approach $f$. Unless otherwise noted, we always repeat a study about $f$ and $\theta$ on all datasets in the leave-one-out manner. More precisely, each score reported in ablation studies is obtained as follows
\begin{equation}\label{eq:score}
 score(\theta, f) = {1\over \|\mathbb{D}\|}\sum_{x\in \mathbb{X}}\left\{\sum_{(\mathbf{D}, \mathbf{B})\in x} {m(\hat{\mathbf{B}}^x_\theta, \mathbf{B}) \over \|x\| } \right\}
\end{equation}
where $m(\cdot)$ indicates a binarization metric, \eg FM; and $\hat{\mathbf{B}}^x_\theta=f^{\mathbb{X}-x}_{\theta}(\mathbf{D})$ indicates the predicted binarized result for a given image $\mathbf{D}$ using the solution $f^{\mathbb{X}-x}_{\theta}$ that is trained on dataset ${\mathbb{X}-x}$ using the setting $\theta$. More precisely, the inner summation of Eq.~\eqref{eq:score} represents the average score for the model $f_{\theta}^{\mathbb{X}-x}$ over all testing samples in $x$, and that the outer summation of Eq.~\eqref{eq:score} further aggregates all leave-one-out average scores, and thus leaves the resulting score only dependent on the used method $f$ with setting $\theta$.

\subsubsection{Does \sauvola{} With Learnable Parameters Work Better?}
\begin{table}[htpb]
    \centering\scriptsize
    \caption{Trainable v.s. non-trainable \sauvola{}.}
    \setlength{\tabcolsep}{2pt}
    \begin{tabular}{@{}c|c|c|l|l|r|r|r|r@{}}
        \hline\hline
        \textbf{WinSize}& \multicolumn{2}{c|}{\textbf{Train?}} &\multicolumn{2}{c|}{\textbf{Converged Value}} & \multicolumn{4}{c}{{\textbf{Binarization Scores}}}\\ \cline{2-9}
        $w$ & \,\,\,$k$\,\,\, & $r$ & \multicolumn{1}{c|}{$k$} & \multicolumn{1}{c|}{$r$}  &\textbf{FM}(\%)$\uparrow$ & \textbf{$F_{ps}$}(\%)$\uparrow$ & {\scriptsize{\textbf{PSNR}}}(db)$\uparrow$ & \textbf{DRD}$\downarrow$\\\hline
         \multicolumn{9}{c}{\textbf{\opencv Parameter Configuration}} \\
         \hline\multirow{4}{*}{11}
         & \xmark  & \xmark & 0.50 & 0.50 &50.09&59.95&13.44&13.73\\
         & \xmark  & \cmark & 0.50 & 0.23 $\pm$ 0.003 &77.31&81.25&15.86&8.17\\
         & \cmark  & \xmark & 0.20 $\pm$ 0.005 & 0.50 &77.92&85.44&15.99&8.09\\
         & \cmark  & \cmark & 0.25 $\pm$ 0.024 & 0.24 $\pm$ 0.004 &\textbf{80.47}&\textbf{85.49}&\textbf{16.05}&\textbf{8.01}\\
         \hline
         \multicolumn{9}{c}{\textbf{\pythreshold Parameter Configuration}} \\
         \hline
         \multirow{4}{*}{15}
         & \xmark  & \xmark & 0.35 & 0.50 &67.37&76.83&14.84&9.96\\
         & \xmark  & \cmark &0.35&0.26 $\pm$ 0.004&79.23&84.01&15.72&8.71\\
         & \cmark  & \xmark &0.22 $\pm$ 0.011&0.50&79.87&86.31&15.84&8.17\\
         & \cmark  & \cmark &0.29 $\pm$ 0.027&0.28 $\pm$ 0.009&\textbf{81.40}&\textbf{86.41}&\textbf{16.35}&\textbf{7.50}\\
         \hline
         \multicolumn{9}{c}{\textbf{\skimage Parameter Configuration}} \\
         \hline
         \multirow{4}{*}{15}
         & \xmark  & \xmark & 0.20 & 0.50 &77.23&85.15&15.55&8.92\\
         & \xmark  & \cmark&0.20&0.25 $\pm$ 0.003&79.51&85.34&15.67&8.43\\
         & \cmark  & \xmark&0.22 $\pm$ 0.008&0.50&79.94&86.37&15.92&8.10\\
         & \cmark  & \cmark&0.29 $\pm$ 0.023&0.28 $\pm$ 0.007&\textbf{81.46}&\textbf{86.47}&\textbf{16.38}&\textbf{7.46}\\
         \hline\hline
    \end{tabular}
    \label{tab:Sauvolaparameters}
\end{table}
Before discussing \snet{}, one must-answer question is whether or not re-implement the classic \sauvola{} algorithm as an algorithmic DNN layer is the right choice, or equivalently, whether or not \sauvola{} hyper-parameters learned from data could generalize better in practice. If not, we should leverage on existing heuristic \sauvola{} parameter settings and use them in \snet{} as non-trainable. 

To answer the question, we start from one set of \sauvola{} hyper-parameters, \ie $\theta=\{w, k, r\}$, and evaluate the corresponding performance of single window \sauvola{}, \ie $g_{\textrm{Sauvola} | \theta}$ under four different conditions, namely, 1) non-trainable $k$ and $r$; 2) non-trainable $k$ but trainable $r$; 3) trainable $k$ but non-trainable $r$; and 4) trainable $k$ and $r$. We further repeat the same experiments for three well-known \sauvola{} hyper-parameter settings \texttt{OpenCV}\footnotemark[1]($w$=11, $k$=0.5, $r$=0.5), \texttt{Scikit-Image}\footnotemark[2]($w$=15, $k$=0.2, $r$=0.5) and \texttt{Pythreshold}\footnote{https://github.com/manuelaguadomtz/pythreshold/blob/master/pythreshold/} ($w$=15, $k$=0.35, $r$=0.5).\par
Table~\ref{tab:Sauvolaparameters} summarizes the performance scores for single-window \sauvola{} with different parameter settings. Each row is about one $score(\theta, f_{\textrm{Sauvola}})$, and the three mega rows represent the three initial $\theta$ settings. As one can see, three prominent trends are: 1) the heuristic \sauvola{} hyper-parameters (\ie the non-trainable $k$ and $r$ setting) from the three open-sourced libraries don't work well for DIBCO-like dataset; 2) allowing trainable $k$ or $r$ leads to better performance, and allowing both trainable gives even better performance; 3) the converged values of trainable $k$ and $r$ are different for different window sizes. We therefore use trainable $k$ and $r$ for each window size in the \texttt{Sauvola} layer (see Sec.~\ref{sec:sauvolaLayer}).

\newcommand{\uniSau}[2]{f^{\textrm{\tiny{Sauvola}}}_{#1 #2}}
\newcommand{\mSau}[2]{f^{\textrm{\tiny{MSauvola}}}_{#1 #2}}

\subsubsection{Does Multiple-Window Help?}
Though it seems that having multiple window sizes for \sauvola{} analysis is beneficial, it is still unclear that 1) how effective it is comparing to a single-window \sauvola{}, and 2) what window sizes should be used. We, therefore, conduct ablation studies to answer both questions. 

More precisely, we first conduct the leave-one-out experiments for the single-window \sauvola{} algorithms for different window sizes with trainable $k$ and $r$. The resulting $score(w, f_\textrm{Sauvola})$ are presented in the upper-half of Table~\ref{tab:uni_win}. Comparing to the best heuristic \sauvola{} performance attained by \texttt{Scikit-Image} in Table~\ref{tab:Sauvolaparameters}, these results again confirm that \sauvola{} with trainable $k$ and $r$ works much better. Furthermore, it is clear that $f_\textrm{Sauvola}$ with different window sizes (except for $w=7$) attain similar scores, possibly because there is no single dominant font size in the 13 public datasets.

\begin{table}[htpb]
    \centering\scriptsize
    \caption{Ablation study on \sauvola{} window sizes}
    \begin{tabular}{c|c|c|c|c|c|c|c|r|r|r|r}
        \hline\hline
        \multicolumn{8}{c|}{\textbf{WinSize}}&\multicolumn{4}{c}{\textbf{Binarization Scores}} \\\hline
        \textbf{7}&\textbf{15}&\textbf{23}&\textbf{31}&\textbf{39}&\textbf{47}&\textbf{55}&\textbf{63}&\textbf{FM (\%)} $\uparrow$&\textbf{$F_{ps}$}~\textbf{(\%)} $\uparrow$&\textbf{ PSNR (db)} $\uparrow$ &\textbf{DRD} $\downarrow$\\\hline\hline
        \multicolumn{12}{c}{\textbf{Single-window Sauvola}}\\\hline

        \cmark&     &     &     &     &     &     &     &77.62&79.60&15.36&9.76\\
             &\cmark&     &     &     &     &     &     &81.47&86.51&16.41&7.47\\
             &     &\cmark&     &     &     &     &     &82.51&87.29&16.43&7.70\\
             &     &     &\cmark&     &     &     &     &82.41&57.09&16.39&7.82\\
             &     &     &     &\cmark&     &     &     &82.23&86.90&16.34&7.93\\
             &     &     &     &     &\cmark&     &     &82.10&86.68&16.30&8.11\\
             &     &     &     &     &     &\cmark&     &82.01&86.54&16.28&8.17\\
             &     &     &     &     &     &     &\cmark&81.92&86.42&16.25&8.25\\\hline
        \multicolumn{12}{c}{\textbf{Multi-window Sauvola}}\\\hline
        
        \cmark&\cmark&     &     &     &     &     &     &81.55&84.41&17.09&6.62\\
        \cmark&\cmark&\cmark&     &     &     &     &     &82.88&85.29&17.32&6.41\\
        \cmark&\cmark&\cmark&\cmark&     &     &     &     &84.71&87.34&17.72&6.04\\
        \cmark&\cmark&\cmark&\cmark&\cmark&     &     &     &87.83&90.70&18.47&5.30\\
        \cmark&\cmark&\cmark&\cmark&\cmark&\cmark&     &     &89.87&92.31&18.87&4.13\\
        \cmark&\cmark&\cmark&\cmark&\cmark&\cmark&\cmark&     &91.36&95.55&19.09&3.73\\
        \cmark&\cmark&\cmark&\cmark&\cmark&\cmark&\cmark&\cmark&\textbf{91.42}&\textbf{95.67}&\textbf{19.15}&\textbf{3.67}\\\hline\hline
    \end{tabular}
    \label{tab:uni_win}
\end{table}

Finally, we conduct the ablation studies of using multiple window sizes in \snet{} in the incremental way, and report the resulting $score(\mathbb{W}, f_\texttt{SauvolaNet})$s in the lower-half of Table~\ref{tab:uni_win}. It is now clear that 1) multi-window does help in \snet{}; and 2) the more window sizes, the better performance scores.
As a result, we use all eight window sizes in \snet{} (see Eq.~\eqref{eq:winsizes}).

\subsection{Comparisons to Classic and SoTA Binarization Approaches}
It is worthy to emphasize that different works~\cite{peng2019document,he2019deepotsu,calvo2019selectional,zhao2019document,howe2013document,lazzara2014efficient} use different protocols for document binarization evaluation. In this section, we mainly follow the evaluation protocol used in \cite{he2019deepotsu}, and its dataset partitions are: 1) training: (H)-DIBCO 2009~\cite{gatos2009icdar}, 2010~\cite{pratikakis2010h}, 2012~\cite{pratikakis2012icfhr}; Bickely-diary dataset~\cite{deng2010binarizationshop}; and Synchromedia Multispectral dataset~\cite{hedjam2015icdar}, and for testing: (H)-DIBCO 2011~\cite{Pratikakis2011icdar}, 2014~\cite{ntirogiannis2014icfhr2014}, and 2016~\cite{pratikakis2016icfhr2016}. We train all approaches using the same evaluation protocol for fairly comparison. As a result, we focus on those recent and open-sourced DNN based methods, and they are SAE~\cite{calvo2019selectional}, DeepOtsu~\cite{he2019deepotsu}, cGANs~\cite{zhao2019document} and MRAtt~\cite{peng2019document}. In addition, heuristic document binarization approaches Otsu~\cite{otsu1979threshold}, Sauvola~\cite{sauvola2000adaptive} and Howe~\cite{howe2013document} are also included. Finally, Sauvola MS~\cite{lazzara2014efficient}, a classic multi-window Sauvola solution is evaluated to better gauge the performance improvement from the heuristic multi-window analysis to the proposed learnable analysis. 

\begin{table}[!t]
    \centering
    \scriptsize
    \caption{Comparison of \texttt{SauvolaNet} and SoTA approaches DIBCO 2011.}
    \definecolor{Gray}{gray}{0.95}
    \begin{tabular}{c|l|r|r|r|r}
        \hline\hline
        \textbf{Dataset}&\textbf{Methods}&\textbf{FM (\%)} $\uparrow$&\textbf{$F_{ps}$} \textbf{(\%)} $\uparrow$&\textbf{ PSNR (db)} $\uparrow$ &\textbf{DRD} $\downarrow$  \\\hline\hline
         \parbox[t]{2mm}{\multirow{10}{*}{\rotatebox[origin=c]{90}{\textbf{DIBCO 2011}}}}
         &Otsu~\cite{otsu1979threshold}&82.10&84.80&15.70&9.00\\
         &Howe~\cite{howe2013document}&91.70&92.00&19.30&3.40\\\cline{2-6}
         &MRAtt~\cite{peng2019document}&93.16&95.23&19.78&2.20\\ 
         &DeepOtsu~\cite{he2019deepotsu}&93.40&95.80&19.90&\textbf{1.90}\\
         &SAE~\cite{calvo2019selectional}&92.77&95.68&19.55&2.52\\
         &DSN~\cite{vo2018binarization}&93.30&\textbf{96.40}&20.10&2.00\\
         &cGANs~\cite{zhao2019document}&93.81&95.26&20.30&1.82\\\cline{2-6}
         &Sauvola~\cite{sauvola2000adaptive}&82.10&87.70&15.60&8.50\\
         &Sauvola MS~\cite{lazzara2014efficient}&79.70&81.78&14.91&11.67\\
         &\snet{}&\textbf{94.32}&\textbf{96.40}&\textbf{20.55}&1.97\\\hline\hline
    \end{tabular}
    \label{tab:compareSoTA2011}
\end{table}

\begin{table}[!t]
    \centering
    \caption{Comparison of \texttt{SauvolaNet} and SoTA approaches in H-DIBCO 2014.}
    \scriptsize
    \definecolor{Gray}{gray}{0.95}
    \begin{tabular}{c|l|r|r|r|r}
        \hline\hline
        \textbf{Dataset}&\textbf{Methods}&\textbf{FM (\%)} $\uparrow$&\textbf{$F_{ps}$} \textbf{(\%)} $\uparrow$&\textbf{ PSNR (db)} $\uparrow$ &\textbf{DRD} $\downarrow$  \\\hline\hline
         \parbox[t]{2mm}{\multirow{11}{*}{\rotatebox[origin=c]{90}{\textbf{H-DIBCO 2014}}}}
         &Otsu~\cite{otsu1979threshold}&91.70&95.70&18.70&2.70\\
         &Howe~\cite{howe2013document}&96.50&97.40&22.20&1.10\\\cline{2-6}
         &MRAtt~\cite{peng2019document}&94.90&95.98&21.09&1.85\\ 
         &DeepOtsu~\cite{he2019deepotsu}&95.90&97.20&22.10&0.90\\
         &SAE~\cite{calvo2019selectional}&95.81&96.78&21.26&1.00\\
         &DSN~\cite{vo2018binarization}&96.70&97.60&23.20&0.70\\
         &DD-GAN~\cite{de2020document}&96.27&97.66&22.60&1.27\\
         &cGANs~\cite{zhao2019document}&96.41&97.55&22.12&1.07\\\cline{2-6}
         &Sauvola~\cite{sauvola2000adaptive}&84.70&87.80&17.80&2.60\\
         &Sauvola MS~\cite{lazzara2014efficient}&85.83&86.83&17.81&4.88\\
         &\snet{}&\textbf{97.83}&\textbf{98.74}&\textbf{24.13}&\textbf{0.65}\\\hline\hline
    \end{tabular}

    \label{tab:compareSoTA2014}
\end{table}

\begin{table}[!t]
    \centering
    \caption{Comparison of \texttt{SauvolaNet} and SoTA approaches DIBCO 2016.}
    \scriptsize
    \definecolor{Gray}{gray}{0.95}
    \begin{tabular}{c|l|r|r|r|r}
        \hline\hline
        \textbf{Dataset}&\textbf{Methods}&\textbf{FM (\%)} $\uparrow$&\textbf{$F_{ps}$} \textbf{(\%)} $\uparrow$&\textbf{ PSNR (db)} $\uparrow$ &\textbf{DRD} $\downarrow$  \\\hline\hline
         \parbox[t]{2mm}{\multirow{11}{*}{\rotatebox[origin=c]{90}{\textbf{DIBCO 2016}}}}
         &Otsu~\cite{otsu1979threshold}&86.60&89.90&17.80&5.60\\ 
         &Howe~\cite{howe2013document}&87.50&82.30&18.10&5.40\\\cline{2-6}
         &MRAtt~\cite{peng2019document}&\textbf{91.68}&94.71&19.59&2.93\\ 
         &DeepOtsu~\cite{he2019deepotsu}&91.40&94.30&19.60&2.90\\
         &SAE~\cite{calvo2019selectional}&90.72&92.62&18.79&3.28\\
         &DSN~\cite{vo2018binarization}&90.10&83.60&19.00&3.50\\
         &DD-GAN~\cite{de2020document}&89.98&85.23&18.83&3.61\\
         &cGANs~\cite{zhao2019document}&91.66&94.58&\textbf{19.64}&\textbf{2.82}\\
         \cline{2-6}
         &Sauvola~\cite{sauvola2000adaptive}&84.60&88.40&17.10&6.30\\  
         &Sauvola MS~\cite{lazzara2014efficient}&79.84&81.61&14.76&11.50\\
         &\snet{}&90.25&\textbf{95.26}&18.97&3.51\\\hline\hline
    \end{tabular}
    \label{tab:compareSoTA2016}
\end{table}

\begin{figure}[!h]
    \centering\scriptsize
    \def\lw{0.19}
    \begin{tabular}{c|c|c|c|c}\hline\hline
    \textbf{Original}& \textbf{cGANs}~\cite{zhao2019document} & \textbf{DeepOtsu}~\cite{he2019deepotsu} &  \textbf{MRAtt}~\cite{peng2019document}&\textbf{SauvolaNet} \\\hline\hline
    \includegraphics[width=\lw\linewidth]{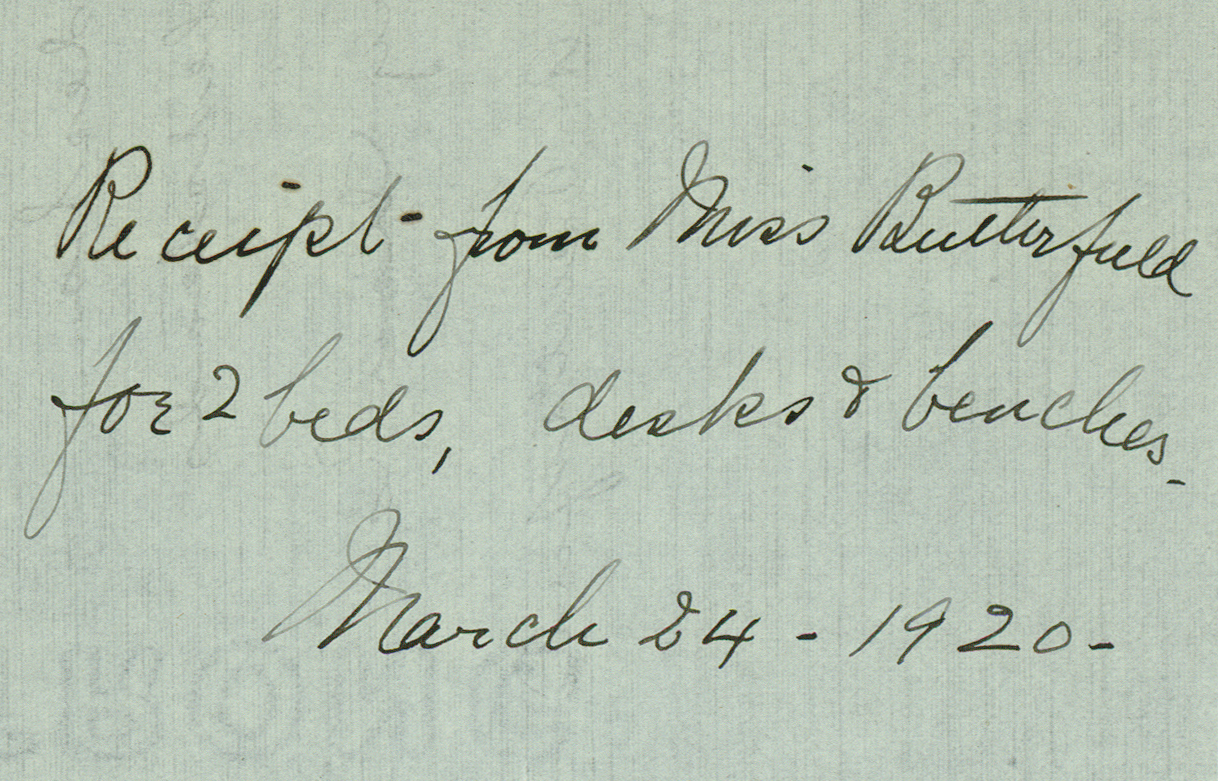}
         & \includegraphics[width=\lw\linewidth]{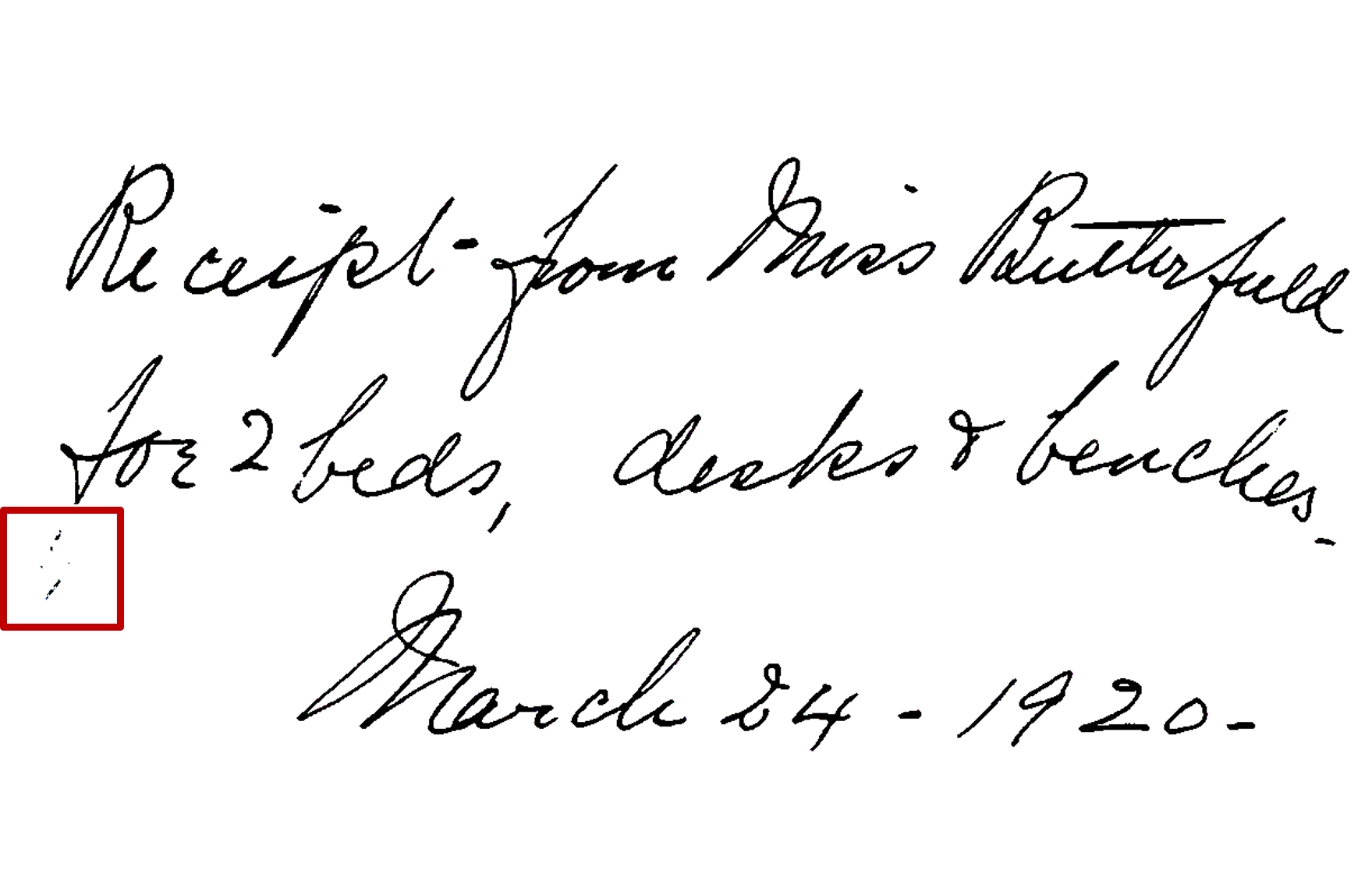}& \includegraphics[width=\lw\linewidth]{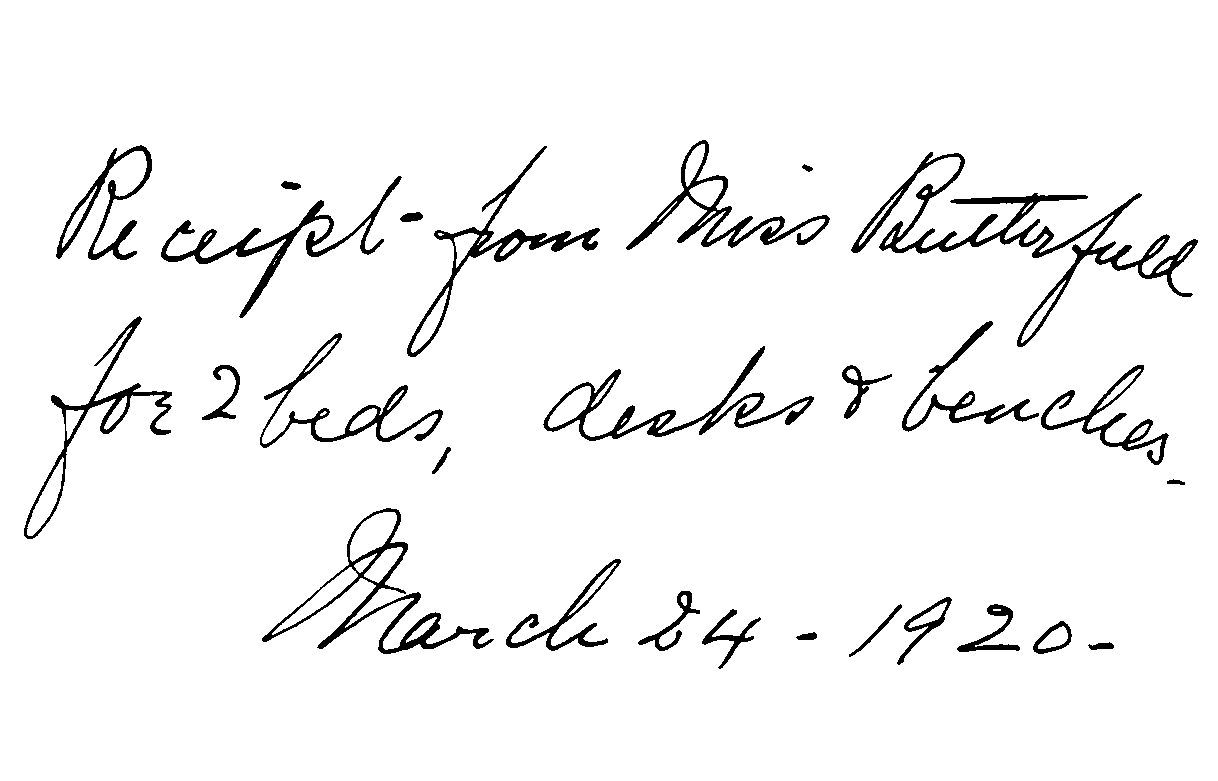}&
         \includegraphics[width=\lw\linewidth]{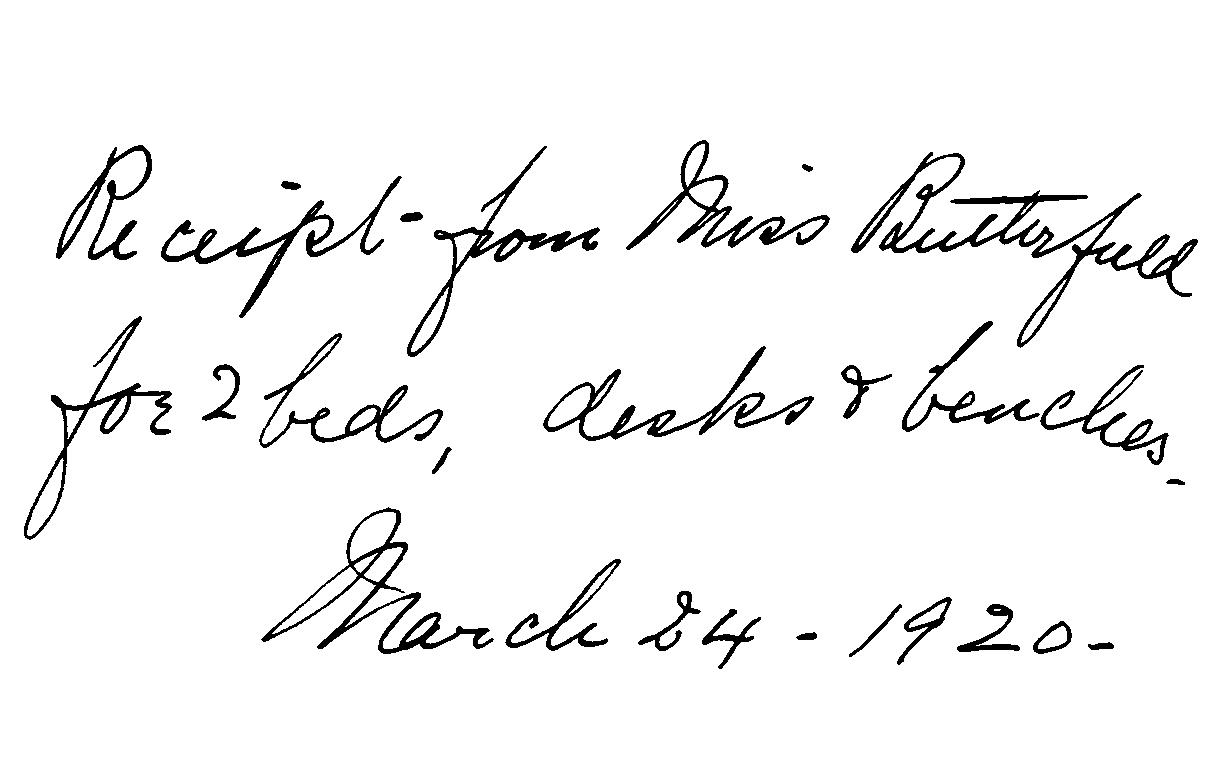}&
         \includegraphics[width=\lw\linewidth]{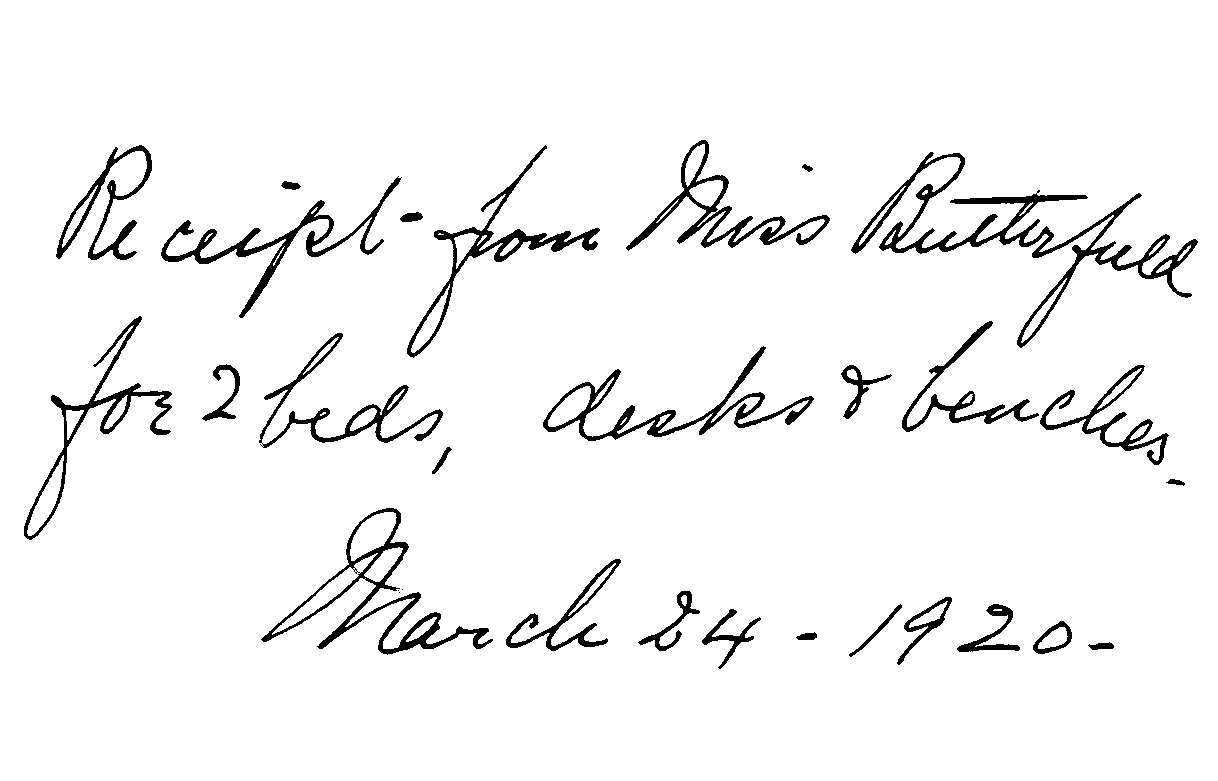}\\
     (a) &95.89  &96.28&95.91&\textbf{97.81} \\\hline
    \includegraphics[width=\lw\linewidth]{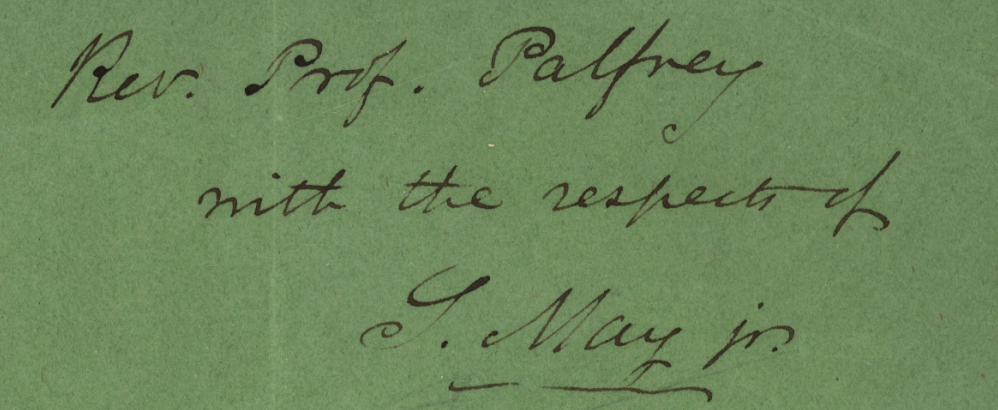}
         & \includegraphics[width=\lw\linewidth]{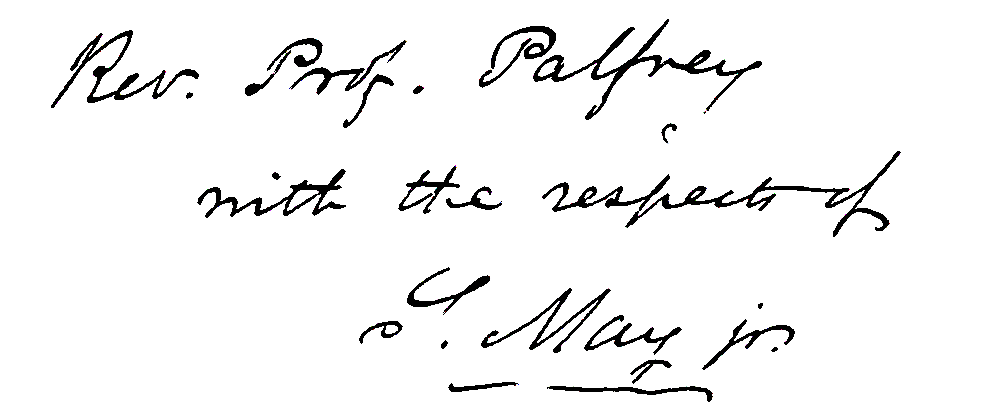}& \includegraphics[width=\lw\linewidth]{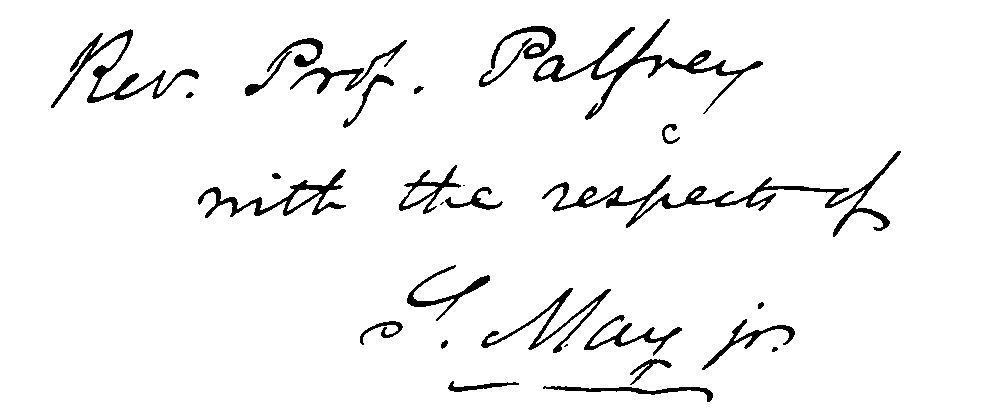}&
         \includegraphics[width=\lw\linewidth]{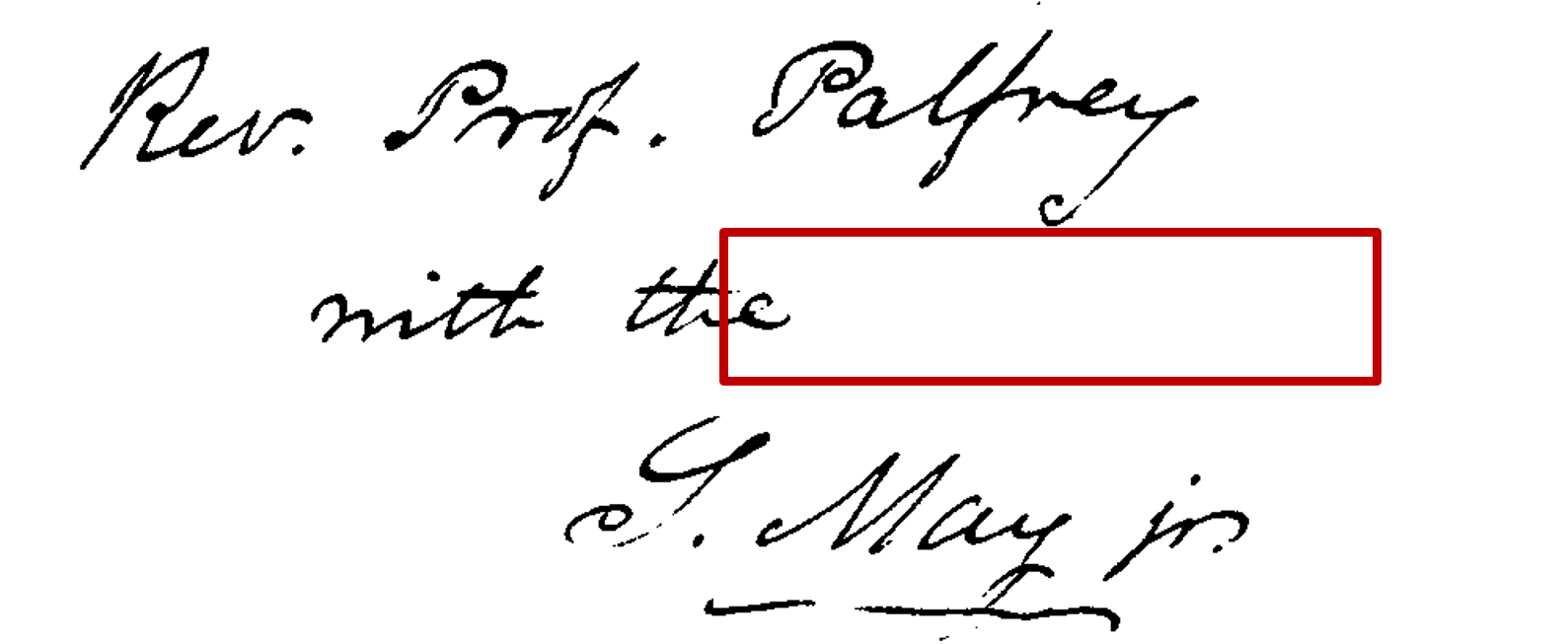}&
         \includegraphics[width=\lw\linewidth]{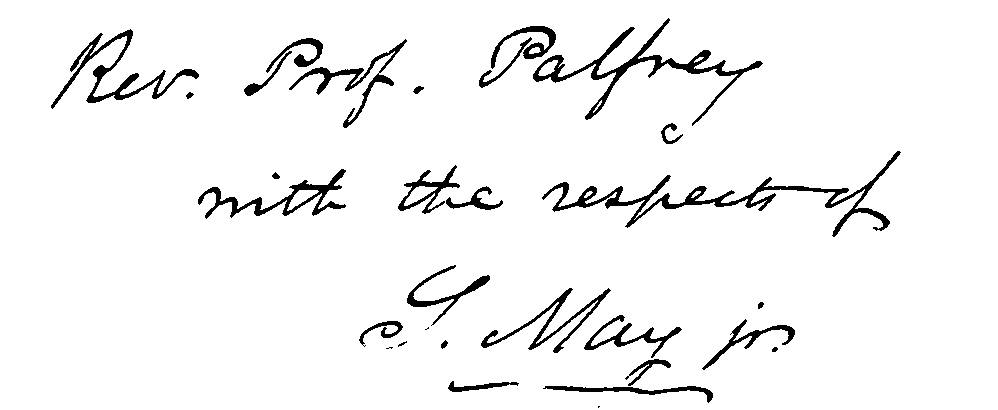}\\
     (b) &93.79  &95.86&86.01&\textbf{96.64} \\\hline
    
     \includegraphics[width=\lw\linewidth]{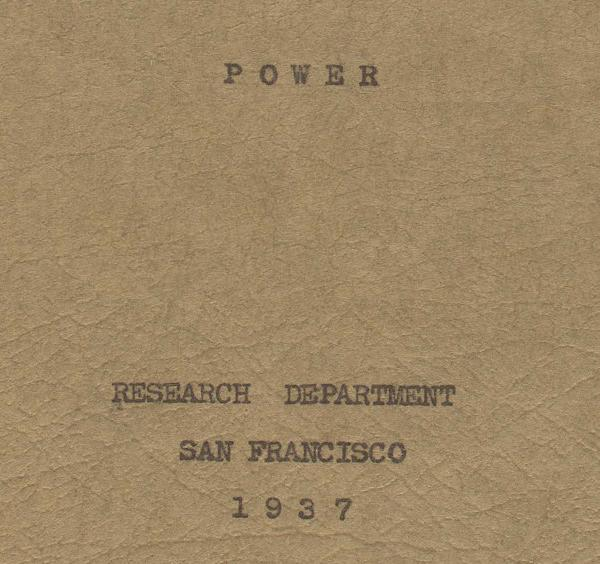}
         & \includegraphics[width=\lw\linewidth]{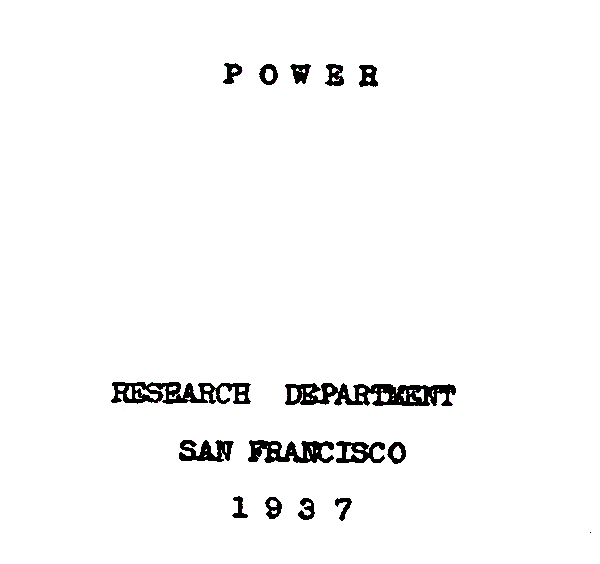}& \includegraphics[width=\lw\linewidth]{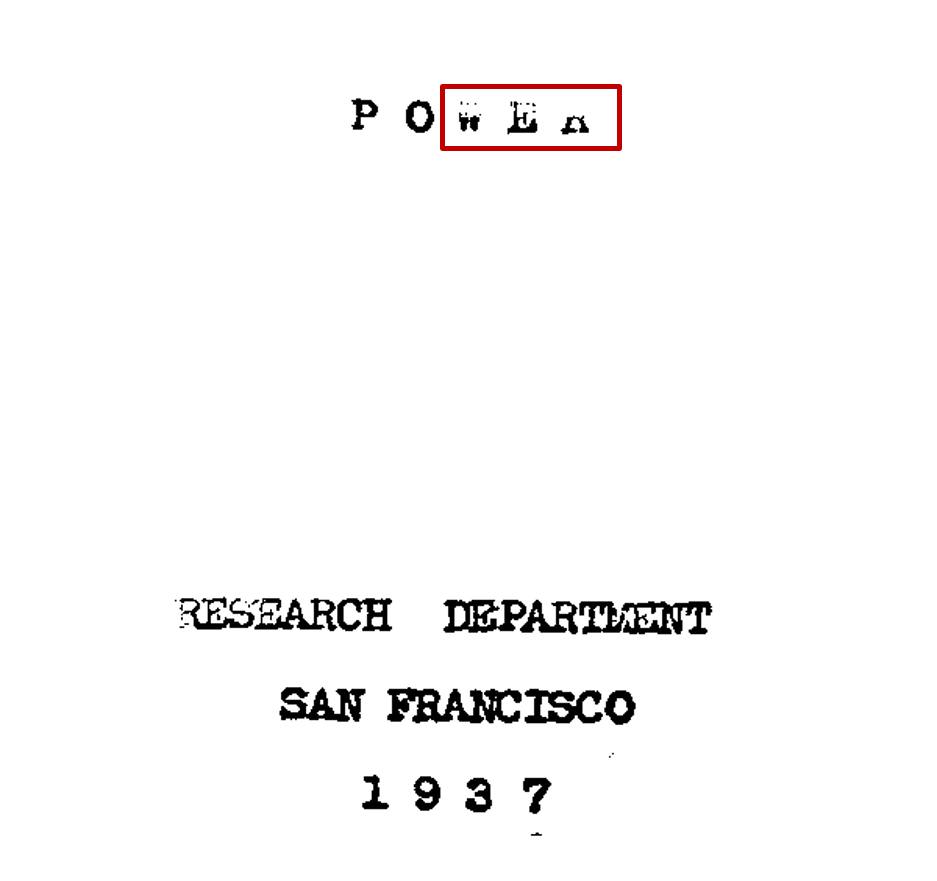}&
         \includegraphics[width=\lw\linewidth]{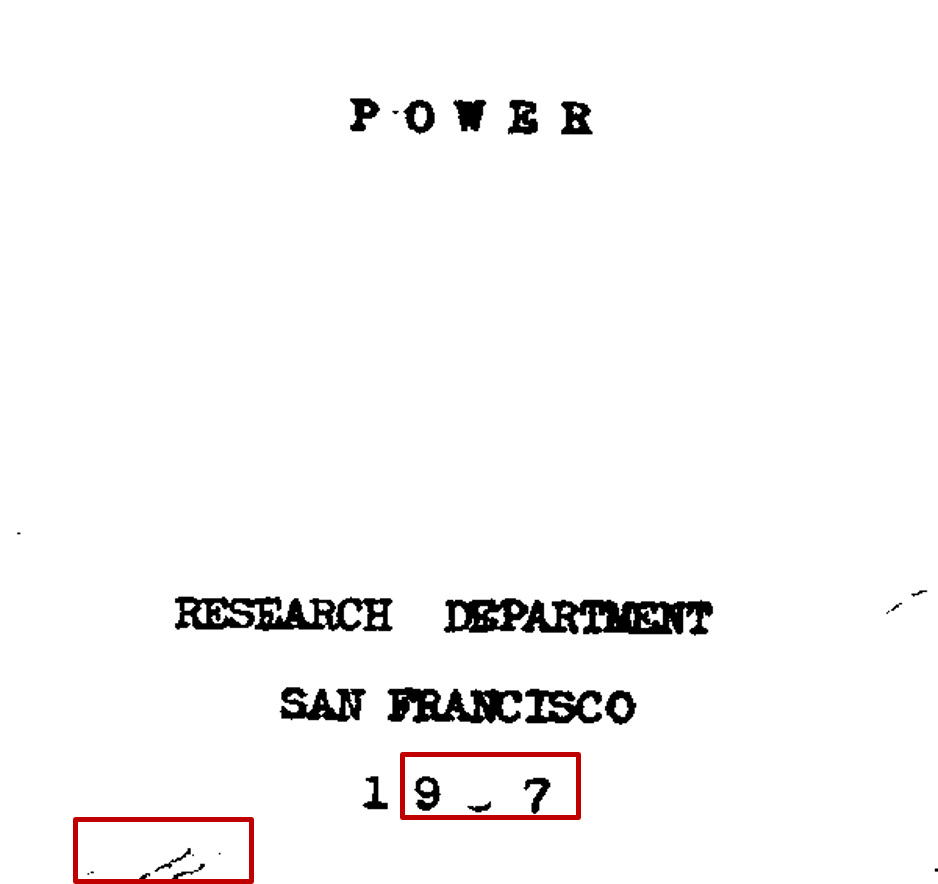}&
         \includegraphics[width=\lw\linewidth]{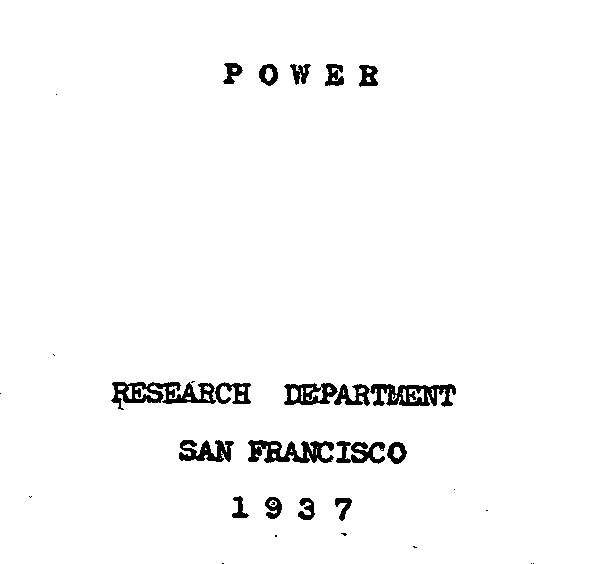}\\
     (c) &91.89  &88.48&90.25&\textbf{93.85} \\\hline
     
    \includegraphics[width=\lw\linewidth]{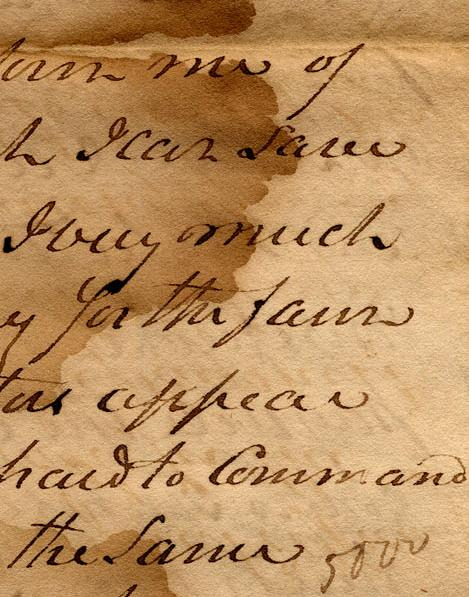}& 
        \includegraphics[width=\lw\linewidth]{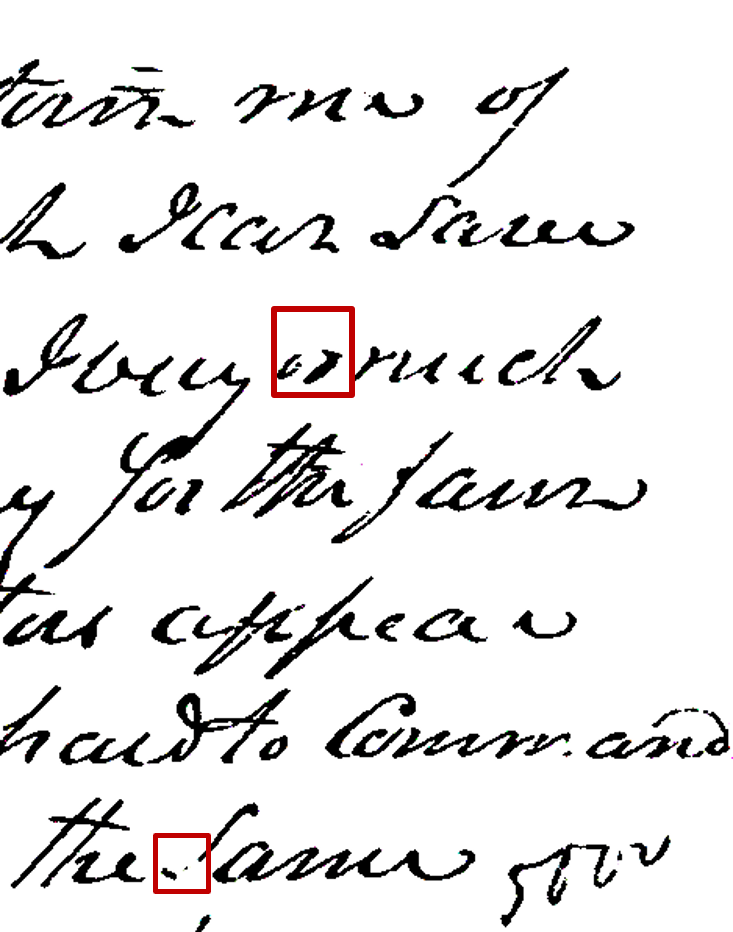}& \includegraphics[width=\lw\linewidth]{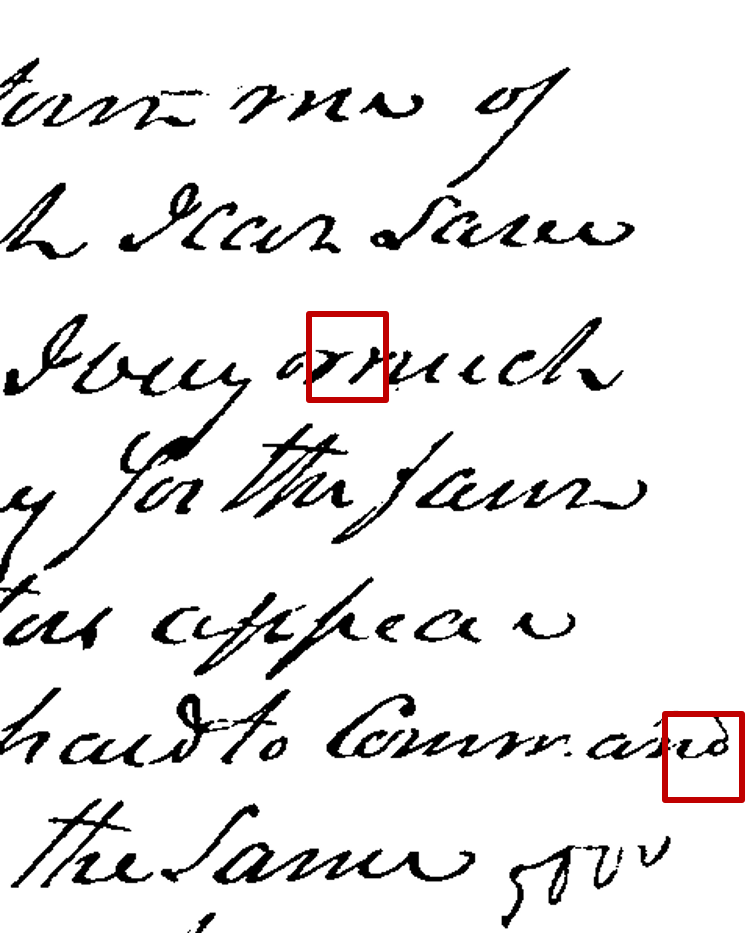}&
         \includegraphics[width=\lw\linewidth]{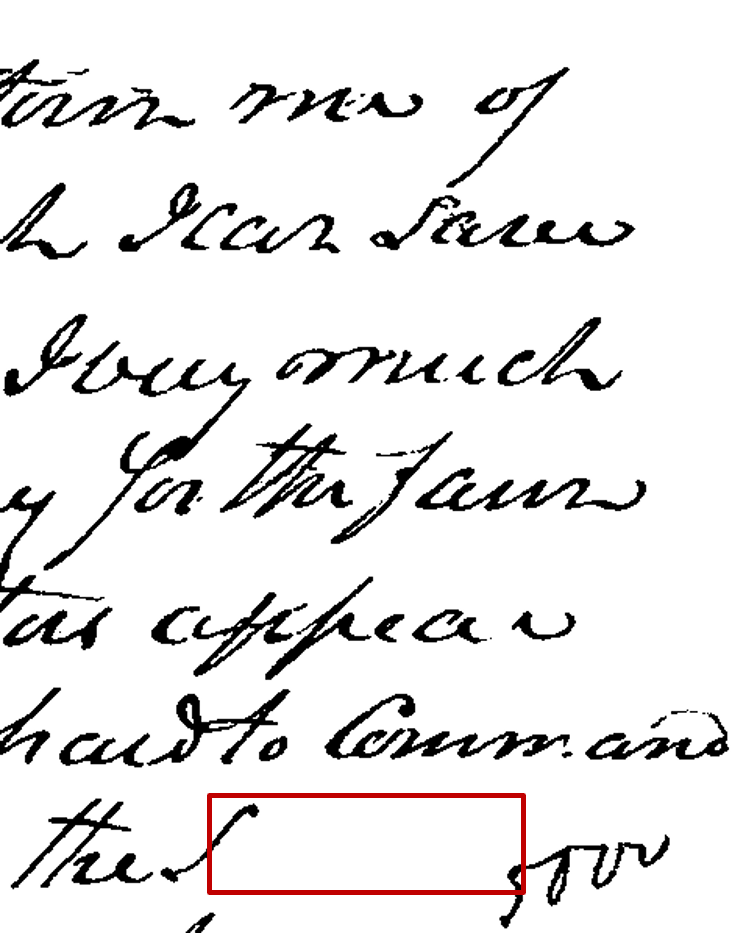}&
         \includegraphics[width=\lw\linewidth]{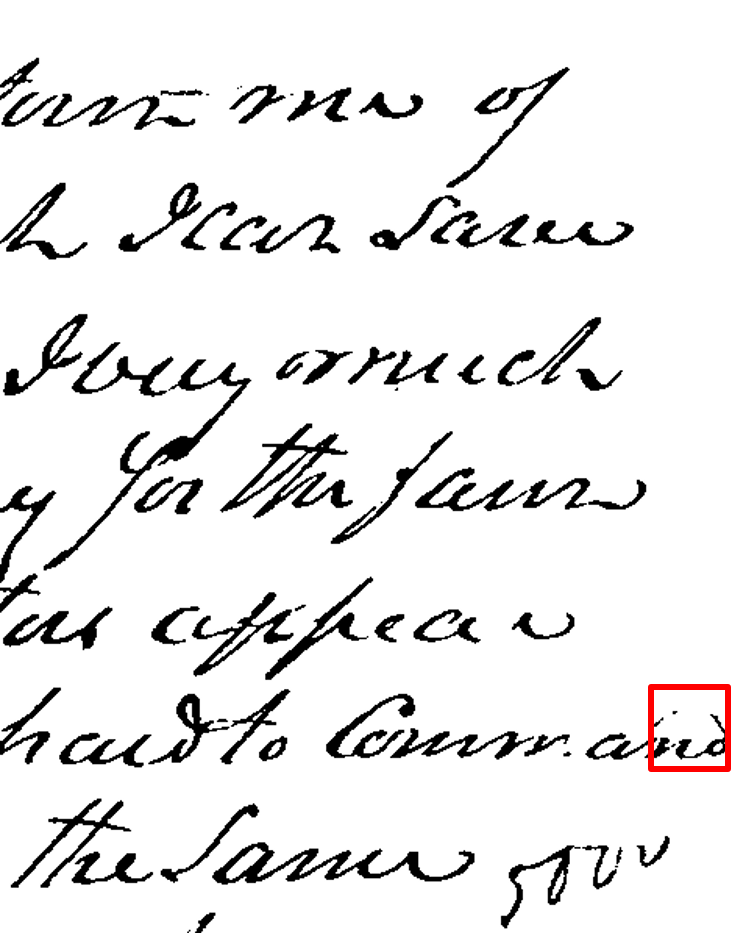}\\
     (d) &91.98  &91.73&91.13&\textbf{94.37}\\\hline\hline
    \end{tabular}
    \caption{Qualitative comparison of \snet{} with SoTA document binarization approaches. Problematic binarization regions are denoted in \colorbox{red}{red boxes}, and the FM score for each binarization result is also included below a result.}
    \label{fig:results}
\end{figure}
Table~\ref{tab:compareSoTA2011}, \ref{tab:compareSoTA2014} and \ref{tab:compareSoTA2016} reports the average performance scores of the four evaluation metrics for all images in each testing dataset. When comparing the three Sauvola based approaches, namely, Sauvola, Sauvola MS, and \snet{}, one may easily notice that the heuristic multi-window solution Sauvola MS does not necessarily outperform the classic Sauvola. However, the \snet{}, again a multi-window solution but with all trainable weights, clearly beat both by large margins for all four evaluation metrics. Moreover, the proposed \snet{} solution outperforms the rest of the classic and SoTA DNN approaches in DIBCO 2011. And \snet{} is comparable to the SoTA solutions in H-DIBCO 2014 and DIBCO 2016. Sample results are shown in Fig.~\ref{fig:results}. More importantly, the \snet{} is super lightweight and only contains 40K parameters. It is much smaller and runs much faster than other DNN solutions as shown in Fig.~\ref{fig:blobPlot}.

\label{sec:experiment}

\section{Conclusion}\label{sec:conclusion}
In this paper, we systematically studied the classic \sauvola{} document binarization algorithm from the deep learning perspective and proposed a multi-window \sauvola{} solution called \snet{}. Our ablation studies showed that the \sauvola{} algorithm with learnable parameters from data significantly outperforms various heuristic parameter settings (see Table~\ref{tab:Sauvolaparameters}). Furthermore, we proposed the \snet{} solution, a \sauvola{}-based DNN with all trainable parameters. The experimental result confirmed that this end-to-end solution attains consistently better binarization performance than non-trainable ones, and that the multi-window \sauvola{} idea works even better in the DNN context with the help of attention (see Table~\ref{tab:uni_win}). Finally, we compared the proposed \snet{} with the SoTA methods on three public document binarization datasets. The result showed that \snet{} has achieved or surpassed the SoTA performance while using a significantly fewer number of parameters (1\% of \texttt{MobileNetV2}) and running at least 5x faster than SoTA DNN-based approaches.

%
%

\bibliographystyle{splncs04}

\end{document}